\documentclass[10pt,twocolumn,letterpaper]{article}

\usepackage{wacv}              

\usepackage{graphicx}
\usepackage{amsmath}
\usepackage{amssymb}
\usepackage{booktabs}
\usepackage{multirow}
\usepackage[accsupp]{axessibility}

\usepackage{cuted}
\usepackage{amsmath,amsfonts,bm}

\newcommand{\fref}[1]{Figure~\ref{#1}}
\newcommand{\eref}[1]{Eq.~(\ref{#1})}
\newcommand{\tref}[1]{Table~\ref{#1}}
\newcommand{\sref}[1]{$\S$~\ref{#1}}
\newcommand{\aref}[1]{Appendix~\ref{#1}}

\def\metfaces{\textsc{MetFaces}}

\def\Pt{\mathbf{P}_{t}}
\def\Dt{\mathbf{D}_{t}}
\def\vepsilon{\bm{\epsilon}}
\def\epsilont{\vepsilon^{\theta}_{t}}

\def\deltah{\Delta\vh}
\def\tildeh{\tilde{\vh}}

\def\tedit{t_\mathrm{edit}}

\def\tboost{t_\mathrm{boost}}

\def\xT{{\bm{x}_T}}

\def\vh{{\bm{h}}}
\def\vg{{\bm{g}}}
\def\xTone{{\bm{x}_T^{(1)}}}
\def\xTtwo{{\bm{x}_T^{(2)}}}

\def\Istyle{{I^{original}}}
\def\Icon{{I^{content}}}

\def\vht{{\bm{h}_t}}

\def\vhttwo{{\bm{h}_t^{(2)}}}

\def\Ione{{I^{(1)}}}
\def\Itwo{{I^{(2)}}}
\def\vhtcon{{\bm{h}_t^{content}}}

\def\vx{{\bm{x}}}
\def\tvx{{\tilde{\bm{x}}}}

\def\vz{{\bm{z}}}
\def\deltah{\Delta\vh}

\def\vepsilon{\bm{\epsilon}}
\def\epsilont{\vepsilon^{\theta}_{t}}

\def\warigari{latent calibration}
\def\Warigari{Latent calibration}

\def\ours{InjectFusion}

\def\thspace{\textit{$\vh$-space}}

\usepackage[linesnumbered,ruled,vlined]{algorithm2e}
\usepackage[pagebackref,breaklinks,colorlinks]{hyperref}

\usepackage[capitalize]{cleveref}
\crefname{section}{Sec.}{Secs.}
\Crefname{section}{Section}{Sections}
\Crefname{table}{Table}{Tables}
\crefname{table}{Tab.}{Tabs.}

\def\wacvPaperID{1288} 
\def\confName{WACV}
\def\confYear{2024}

\begin{document}

\title{Training-free Content Injection using h-space in Diffusion Models}

\author{Jaeseok Jeong$^*$\\
\and
Mingi Kwon$^*$\\
Yonsei University\\
Seoul, Republic of Korea\\
{\tt\small \{jete\_jeong,kwonmingi,yj.uh\}@yonsei.ac.kr}
\and
Youngjung Uh$^\dag$\\
\vspace{-2.0em}
}

\maketitle
\def\thefootnote{*}\footnotetext{These authors contributed equally to this work}
\def\thefootnote{$\dag$}\footnotetext{corresponding author}
    

\begin{abstract}
Diffusion models (DMs) synthesize high-quality images in various domains.
However, controlling their generative process is still hazy because the intermediate variables in the process are not rigorously studied. 
Recently, the bottleneck feature of the U-Net, namely \thspace{}, is found to convey the semantics of the resulting image. It enables StyleCLIP-like latent editing within DMs.
In this paper, we explore further usage of \thspace{} beyond attribute editing, and introduce a method to inject the content of one image into another image by combining their features in the generative processes. Briefly, given the original generative process of the other image, 1) we gradually blend the bottleneck feature of the content with proper normalization, and 2) we calibrate the skip connections to match the injected content.
Unlike custom-diffusion approaches, our method does not require time-consuming optimization or fine-tuning. Instead, our method manipulates intermediate features within a feed-forward generative process. Furthermore, our method does not require supervision from external networks. \href{https://curryjung.github.io/InjectFusion/}{Project Page}
\end{abstract}


\vspace{-1.5em}
\section{Introduction}
\label{sec:intro}
Diffusion models (DMs) have gained recognition in various domains due to their remarkable performance in random generation~\cite{ho2020denoising,song2020denoising}. Naturally, researchers and practitioners seek ways to control the generative process. In this sense, text-to-image DMs provide a way to reflect a given text for generating diverse images using classifier-free guidance~\cite{nichol2021glide,ramesh2022hierarchical,saharia2022photorealistic,rombach2022high,balaji2022ediffi,gafni2022make}. In the same context, image guidance synthesizes random images that resemble the reference images that are given for the guidance~\cite{choi2021ilvr,avrahami2022blended,lugmayr2022repaint,meng2021sdedit,chung2022improving}. 
On the other hand, deterministic DMs, such as ODE samplers, have been used to edit real images while preserving most of the original image~\cite{song2020denoising,song2020score,jolicoeur2021gotta,liu2022pseudo,lu2022dpm}. DiffusionCLIP~\cite{kim2021diffusionclip} and Imagic~\cite{kawar2022imagic} first embed an input image into noise and finetune DMs for editing. 
While these approaches provide some control for DMs,  the intermediate variables in the process are not rigorously studied, as opposed to the latent space of generative adversarial networks (GANs).
Critically, previous studies do not provide insight into the intermediate features of DMs.

\begin{figure}[!t]
    \centering
    \includegraphics[width=0.85\linewidth]{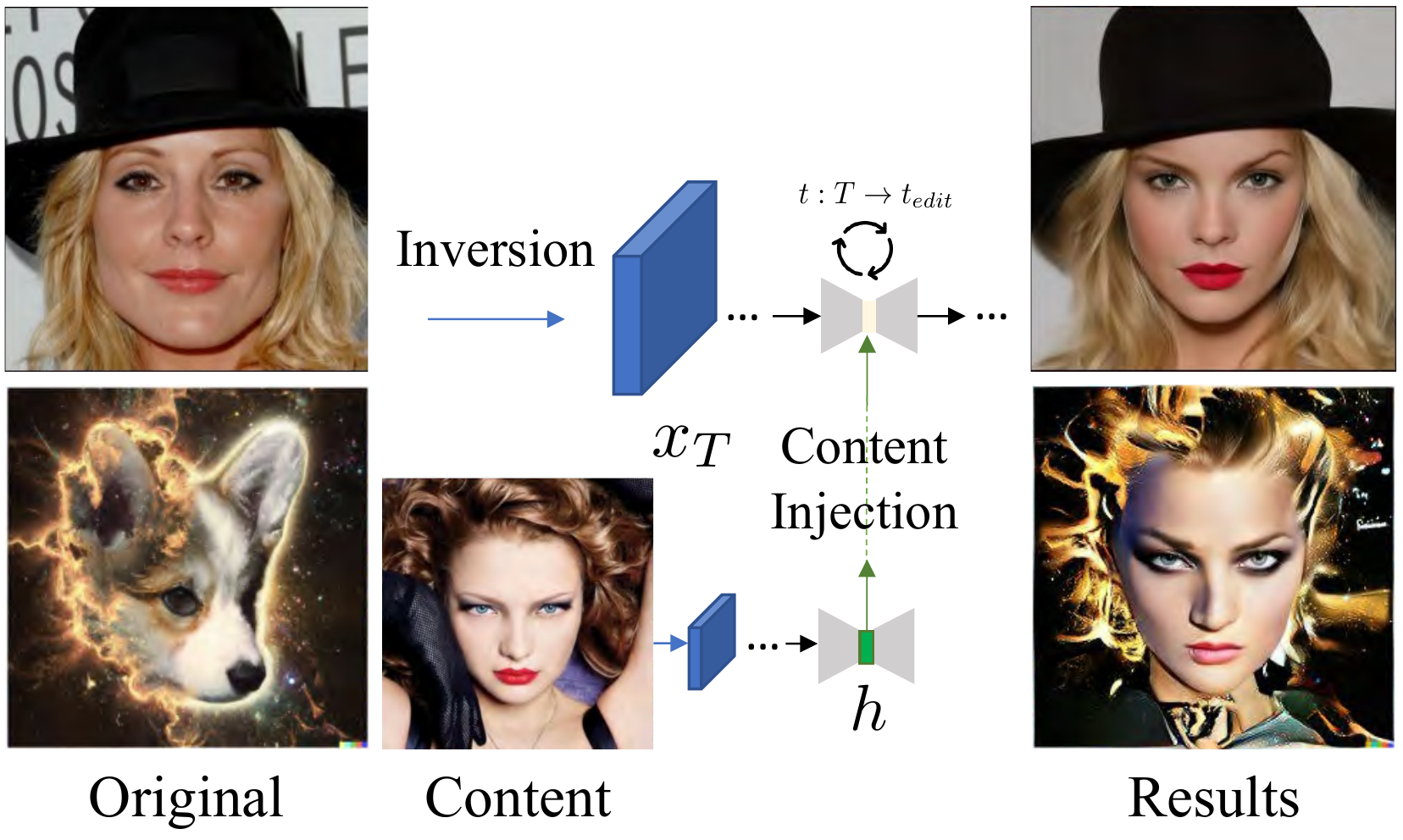}
    \vspace{-0.5em}
    \caption{\textbf{Overview of \ours{}.} 
    During the content injection, the bottleneck feature map is recursively injected during the sampling process started from the inverted $\vx_T$ of images. The target content is reflected in the result images while preserving the original images.}
    \vspace{-1.5em}
    \label{fig:overview}
\end{figure}

Recently, Asyrp~\cite{kwon2022diffusion} discovered a hidden latent space of pretrained DMs located at the bottleneck of the U-Net, named \thspace{}. Shifting the latent feature maps along a certain direction enables semantic attribute changes, such as adding a smile. When combined with deterministic inversion, it allows real image manipulation using a pretrained frozen DM. However, its application is limited to changing certain attributes, and it does not provide as explicit operations as in GANs, such as replacing feature maps.

In this paper, we explore further usage of \thspace{} beyond attribute editing and introduce a method that injects the content of one image into another image.
\fref{fig:overview} overviews our new generative process for content injection. It starts by inverting two images into noises. Instead of running generative processes from them individually, we set one generative process as an original and inject the bottleneck features of the other generative process. As the bottleneck features convey the semantics of the resulting image, it is equivalent to injecting the content. The injection happens recursively along the timesteps.

However, unlike GAN, DMs are usually designed with U-Net which has skip connections. If one directly changes the bottleneck only, it distorts the relation between the skip connection and the bottleneck. Our method, named InjectFusion, treats this problem with two methods. 
1) \ours{} blends the content bottleneck to the original bottleneck gradually along the generative process. The blended feature is properly normalized to keep the correlation with the skip connections.
2) \ours{} calibrates the latent $\vx_t$ directly to preserve the correlation between \thspace{} and skip connections.
This calibration is not only able to be used for InjectFusion but also for any other feature manipulation methods.

InjectFusion enables content injection using pretrained unconditional diffusion models without any training. To the best of our knowledge, our method is the first to tackle these applications without additional training or extra networks. It provides convenience for users to experiment with existing pretrained DMs. In the experiments, we analyze the effect of individual components and demonstrate diverse use cases. Although there is no comparable method with a perfect fit, we compare InjectFusion against closely related methods, including DiffuseIT \cite{kwon2022diffuseIT}.

\section{Background}
In this section, we review various approaches for controlling the results of DMs and cover preliminaries.
\subsection{Diffusion models and controllability}

After DDPMs \cite{ho2020denoising} provide a universal approach for DMs, Song et al. \cite{song2020score} unify DMs with score-based models in SDEs. Subsequent works have focused on improving generative performance of DMs ~\cite{nichol2021improved,karras2022elucidating,choi2022perception,song2020denoising,watson2022learning}. Other works attempt to manipulate the resulting images by replacing latent variables in DMs and generating random images with the color or strokes of the desired images \cite{choi2021ilvr,meng2021sdedit} but they fall short of content injection.

Recently, some works have proposed to control DMs by manipulating latent features in DMs. Asyrp \cite{kwon2022diffusion} considers the bottleneck of U-Net as a semantic latent space (\thspace{}) through the asymmetric reverse process. However, it focuses only on semantic editing, e.g., making a person smile. Plug-and-Play \cite{tumanyan2023plug} injects an intermediate feature in DMs to provide structural guidance. However, it does not consider the correlation between the skip connection and the feature. Similarly, injecting self-attention features enables semantic image editing by retaining structure or objects/characters \cite{tumanyan2023plug, cao2023masactrl}. However, they should rely on text prompts to determine the destinations, which is often vague and insufficient in describing abstract and fine-grained visual concepts.

ADM \cite{dhariwal2021diffusion} introduces gradient-guidance to control generative process~\cite{sehwag2022generating,avrahami2022blended,liu2021more,nichol2021glide}, but it does not allow detailed manipulation. The guidance controls the reverse process of DMs and can be extended to image-guided image translation without extra training but it depends on the external model (e.g. DINO ViT \cite{caron2021emerging}) and struggles to overcome a huge disparity in color distribution. \cite{kwon2022diffuseIT} 

\subsection{Injecting contents from exemplar images}

For given exemplar images with an object, Dreambooth variants~\cite{ruiz2022dreambooth, kumari2023multi} fine-tune pretrained DMs to generate different images containing the object. Instead of fine-tuning the whole model, LoRA variants \cite{lora_repo, zhang2023adding, li2023gligen} introduce auxiliary networks or fine-tune a tiny subset of the model.
As opposed to modifying models, textual inversion variants~\cite{gal2022image, han2023highly} embed visual concepts into text embeddings for the same task.
However, these methods require extra training or optimization steps to reflect the exemplars. On the other hand, our method does not require training or optimization but works on frozen pretrained models.
In addition, while these methods rely on the form of text to reflect the exemplars, our method directly works on the intermediate features in the model.

ControlNet variants \cite{zhang2023adding, mou2023t2i,li2023gligen} can inject structural contents as a condition in the form of an edge map, segmentation mask, pose, and depth map. However, the control is limited to structure and shape. Our method preserves most of the content in the exemplar.

Some works utilize the inversion capability of DMs \cite{hertz2022prompt, brooks2023instructpix2pix, mokady2023null, tumanyan2023plug, cao2023masactrl}, which enables injecting contents during the reconstruction process. However, most of them rely on language to insert the contents.



\subsection{Style transfer}

Recently, neural style transfer \cite{gatys2015neural} has evolved with the advancement of DMs and neural network architecture \cite{dosovitskiy2020image}. Some style transfer methods leverage a style encoder \cite{ruta2023aladin} to enable pretrained DMs to be conditioned on the visual embedding from style reference images \cite{tarres2023parasol,ruta2023diff}. StyleDrop \cite{sohn2023styledrop} achieves outstanding performance in extracting style features from visual examples but how to control content and shape has not been provided. Since it is vision transformer \cite{dosovitskiy2020image}, universal spatial control approach of DMs \cite{zhang2023adding} cannot be adapted 

Exploiting external segmentation mask models and explicit appearance encoder enables decomposing the structure and appearance in \cite{goel2023pair} for style transfer, but it requires training DMs and the encoder from scratch.


\subsection{Denoising Diffusion Implicit Model (DDIM)}
\label{ddim}
Diffusion models learn the distribution of data by estimating denoising score matching with $\epsilont$. In the denoising diffusion probabilistic model (DDPM) \cite{ho2020denoising}, the forward process is defined as a Markov process that diffuses the data through parameterized Gaussian transitions.
DDIM \cite{song2020denoising}  redefines DDPM as $q_{\sigma}(\vx_{t-1}|\vx_{t}, \vx_{0})=\mathcal{N}(\sqrt{\alpha_{t-1}} \vx_{0}+\sqrt{1-\alpha_{t-1}-\sigma_{t}^{2}} \cdot \frac{\vx_{t}-\sqrt{\alpha_{t}} \vx_{0}}{\sqrt{1-\alpha_{t}}}, \sigma_{t}^{2} \boldsymbol{I})$, where $\{\beta_{t}\}^{T}_{t=1}$ is the variance schedule and $\alpha_{t} = \prod^{t}_{s=1} (1-\beta_s)$. 
Accordingly, the reverse process becomes: 
\begin{equation}
\begin{aligned}
\label{ddim_reverse}
    \vx_{t-1}= & \sqrt{\alpha_{t-1}} \underbrace{\left(\frac{\vx_{t}-\sqrt{1-\alpha_{t}} \epsilont\left(\vx_{t}\right)}{\sqrt{\alpha_{t}}}\right)}_{\text {"predicted } \vx_{0} \text { " }} \\
    & +\underbrace{\sqrt{1-\alpha_{t-1}-\sigma_{t}^{2}} \cdot \epsilont\left(\vx_{t}\right)}_{\text {"direction pointing to } \vx_{t} \text { " }}+\underbrace{\sigma_{t} \vz_{t}}_{\text {random noise }},
\end{aligned}
\end{equation}
where $\sigma_t = \eta \sqrt{\left(1-\alpha_{t-1}\right) /\left(1-\alpha_{t}\right)} \sqrt{1-\alpha_{t} / \alpha_{t-1}}$. When $\eta=0$, the process becomes deterministic.

\subsection{Asymmetric reverse process (Asyrp)}
Asyrp \cite{kwon2022diffusion} introduces the asymmetric reverse process for using \thspace{} as a semantic latent space. \thspace{} is the bottleneck of U-Net, which is distinguished from the latent variable $\vx_t$. For real image editing, they invert $\vx_0\sim p_{real}(\vx)$ into $\vx_{T}$ through the DDIM forward process, and generate $\tvx_0$ using the new $\tildeh_t$ in the modified DDIM reverse process. They use an abbreviated version of \eref{ddim_reverse}. We follow the notation of Asyrp throughout this paper:
\begin{equation}
\label{sampling_eq}
    \vx_{t-1} = \sqrt{\alpha_{t-1}}\ \Pt(\epsilont(\vx_t)) + \Dt(\epsilont(\vx_t)) + \sigma_{t} \vz_{t},
\end{equation}
where $\Pt(\epsilont(\vx_t))$ denotes the predicted $\vx_0$ and $\Dt(\epsilont(\vx_t))$ denotes the direction pointing to $\vx_t$. We abbreviate $\Pt(\epsilont(\vx_t))$ as $\Pt$ and $\Dt(\epsilont(\vx_t))$ as $\Dt$ when the context clearly specifies the arguments. Following Asyrp, we omit $\sigma_{t} \vz_{t}$ when $\eta = 0$. 
Then, Asyrp becomes:
\begin{equation}
\label{bm-sampling}
    \tvx_{t-1} = \sqrt{\alpha_{t-1}}\ \Pt(\epsilont(\tvx_t|\tildeh_t)) + \Dt(\epsilont(\tvx_t|\vh_t)) + \sigma_{t} \vz_{t},
\end{equation}
where $\tilde{\vx}_T=\tvx_T$ and then $\epsilont(\tvx_t|\tildeh_t)$ replaces the original U-Net feature maps $\vh_t$ with $\tildeh_t$.
They show that the modification of \thspace{} in both $\Pt$ and $\Dt$ brings a negligible change in the results. Therefore, the key idea of Asyrp is to modify only \thspace{} of $\Pt$ while preserving $\Dt$. 

Quality boosting, introduced by Asyrp, is a stochastic noise injection when the image is almost determined. It enhances fine details and reduces the noise of images while preserving the identity of the image. The whole process of Asyrp is as follows.
\begin{equation}
\begin{aligned}
&\tilde{\vx}_{t-1} = \\
    &\begin{cases}
    &\sqrt{\alpha_{t-1}}\ \Pt(\epsilont(\tilde{\vx}_t|\tildeh_t)) +\Dt \quad \text { if } T \ge t \ge \tedit \\
    &\sqrt{\alpha_{t-1}}\ \Pt(\epsilont(\tilde{\vx}_t|\vh_t)) +\Dt \quad \text { if } \tedit > t  \ge \tboost \\
    &\sqrt{\alpha_{t-1}}\ \Pt(\epsilont(\tilde{\vx}_t|\vh_t)) +\Dt + \sigma_{t}^{2}\vz  \ \text { if } \tboost > t
    \end{cases}
\end{aligned}
\end{equation}

which consists of editing, denoising, and quality boosting intervals where the hyperparameter $\tedit$ determines the editing interval and $\tboost$ determines the quality boosting interval. Following Asyrp, we apply quality boosting to all figures except for ablation studies.

\begin{figure}[!t]
    \centering
    \includegraphics[width=0.9\linewidth]{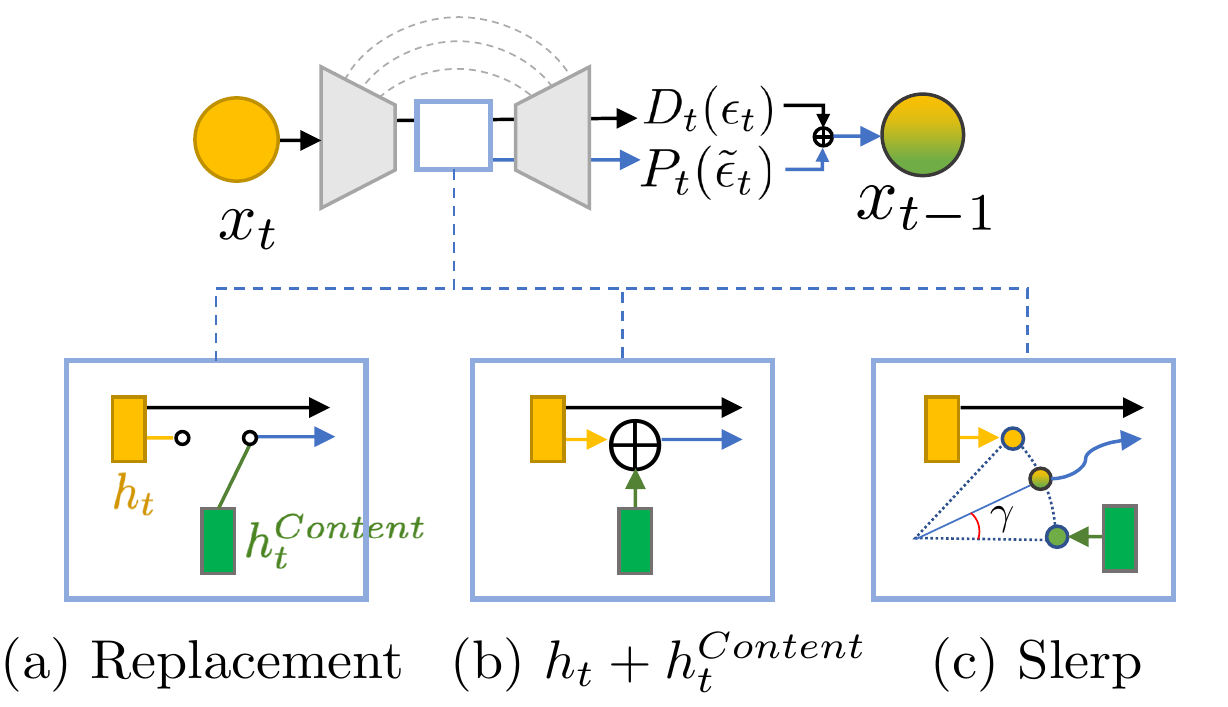}
    \vspace{-0.5em}
    \caption{\textbf{Illustration of content injection methods}. (a) and (b) provide content injection but suffer quality degradation. Compared to them, (c) allows successful content injection by preserving statistics in DMs and gradually increasing the ratio of the target content.}
    \vspace{-0.5em}
    \label{fig:slerp_compare}
\end{figure}

\section{Method}
\label{sec:method}
In this section, we explore the interesting properties of \thspace{} with Asyrp~\cite{kwon2022diffusion} and design a method for content injection. We start by simply replacing $\vh_t$ of one sample with that of another sample and observe its drawbacks in \sref{sec:replace}. Then we introduce an important requirement for mixing two $\vh_t$'s in \sref{sec:slerp}. Furthermore, we propose \warigari{} to retain the crucial elements in \sref{sec:style_calibration}.

\subsection{Role of \textit{\textbf{h-space}}}
\label{sec:replace}
\thspace{}, the deepest bottleneck of the U-Net in the diffusion models (DMs), contains the semantics of the resulting images to some extent. In other words, a change in \thspace{} with Asyrp~\cite{kwon2022diffusion} leads to editing the resulting image. Formally, setting $\tildeh_t = \vh_t + \deltah_t$ for $t\in[T,\tedit]$ modifies the semantics, where $\deltah_t$ is the direction of desired attribute. 
The reverse process becomes $\tilde{\vx}_{t-1} = \sqrt{\alpha_{t-1}}\ \Pt(\epsilont(\tilde{\vx}_t|\tildeh_t)) +\Dt(\epsilont(\tvx_t|\vh_t))$, where $\tildeh_t = \vh_t + \deltah_t^{\text{attr}}$.

We start with a question: Does $\vh$ solely specify the semantics of the resulting image as in the latent codes in GANs? I.e., would replacing $\vh$ totally change the output?

To answer the question, we invert two images $\Ione{}$ and $\Itwo{}$ to noises $\xTone{}$ and $\xTtwo{}$ via forward process, respectively.
Then we replace $\{\vht\}$\footnote{Note that the reverse process is recursive. The reason we denote $\{\vht\}$ instead of $\vht^{(1)}$ is that it differs from $\vht^{(1)}$ after the first replacement.} from $\xTone{}$ with $\{\vhttwo{}\}$ from $\xTtwo{}$ during the reconstruction (i.e., reverse process).
Formally, $\tilde{\vx}_{t-1} = \sqrt{\alpha_{t-1}}\ \Pt(\epsilont(\tilde{\vx}_t|\vhttwo{})) +\Dt(\epsilont(\tvx_t|\vh_t))$, $\tilde{\vx}_{T} = \xTone{}$; which is illustrated in \fref{fig:slerp_compare}a.

\begin{figure}[!t]
    \centering
    \includegraphics[width=1\linewidth]{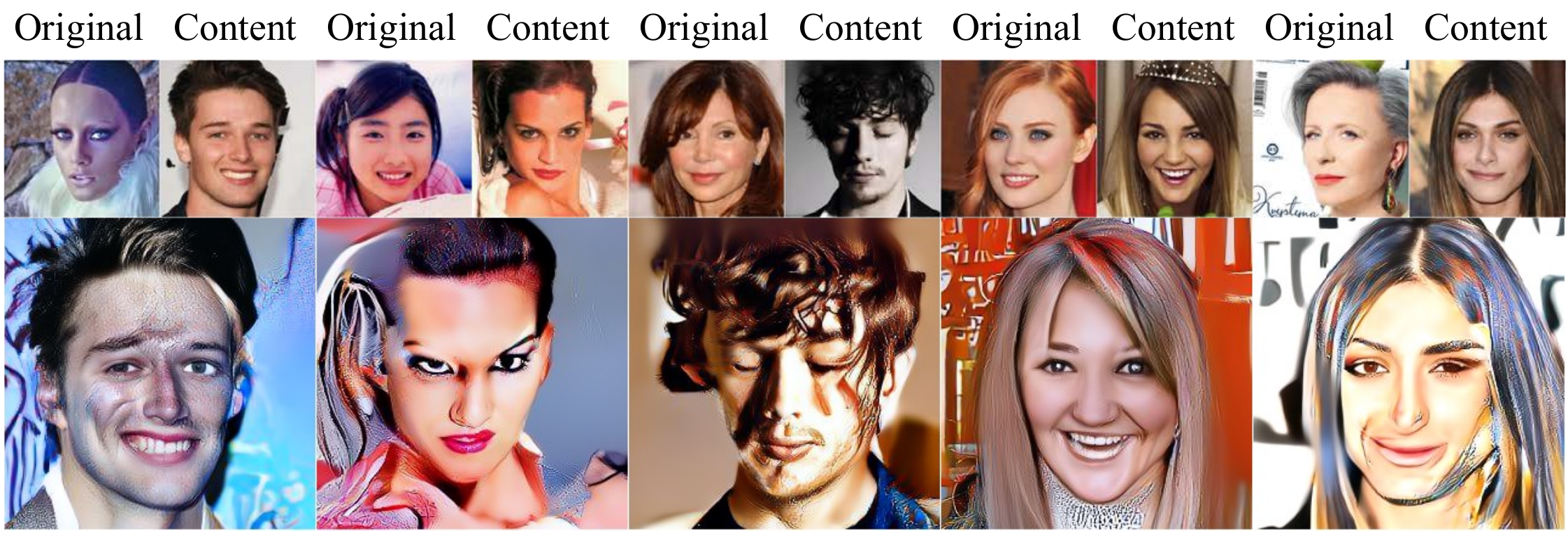}
    \vspace{-1.5em}
    \caption{\textbf{Preliminary experiment.} Na\"ive replacement of $\vh$ somehow combines the content and the original image. However, it severely degrades image quality.}  
    \vspace{-0.5em}
    \label{fig:replace}
\end{figure}

Interestingly, the resulting images with the replacement contain the people in $\Itwo{}$ with some elements of $\Ione{}$ such as color distributions and backgrounds as shown in \fref{fig:replace}.
This phenomenon suggests that the main content is specified by $\vh$ and the other aspects come from the other components, e.g., features in the skip connections. Henceforth, we name $\vhttwo{}$ as $\vhtcon{}$.

\begin{figure}[!t]
    \centering
    \includegraphics[width=\linewidth]{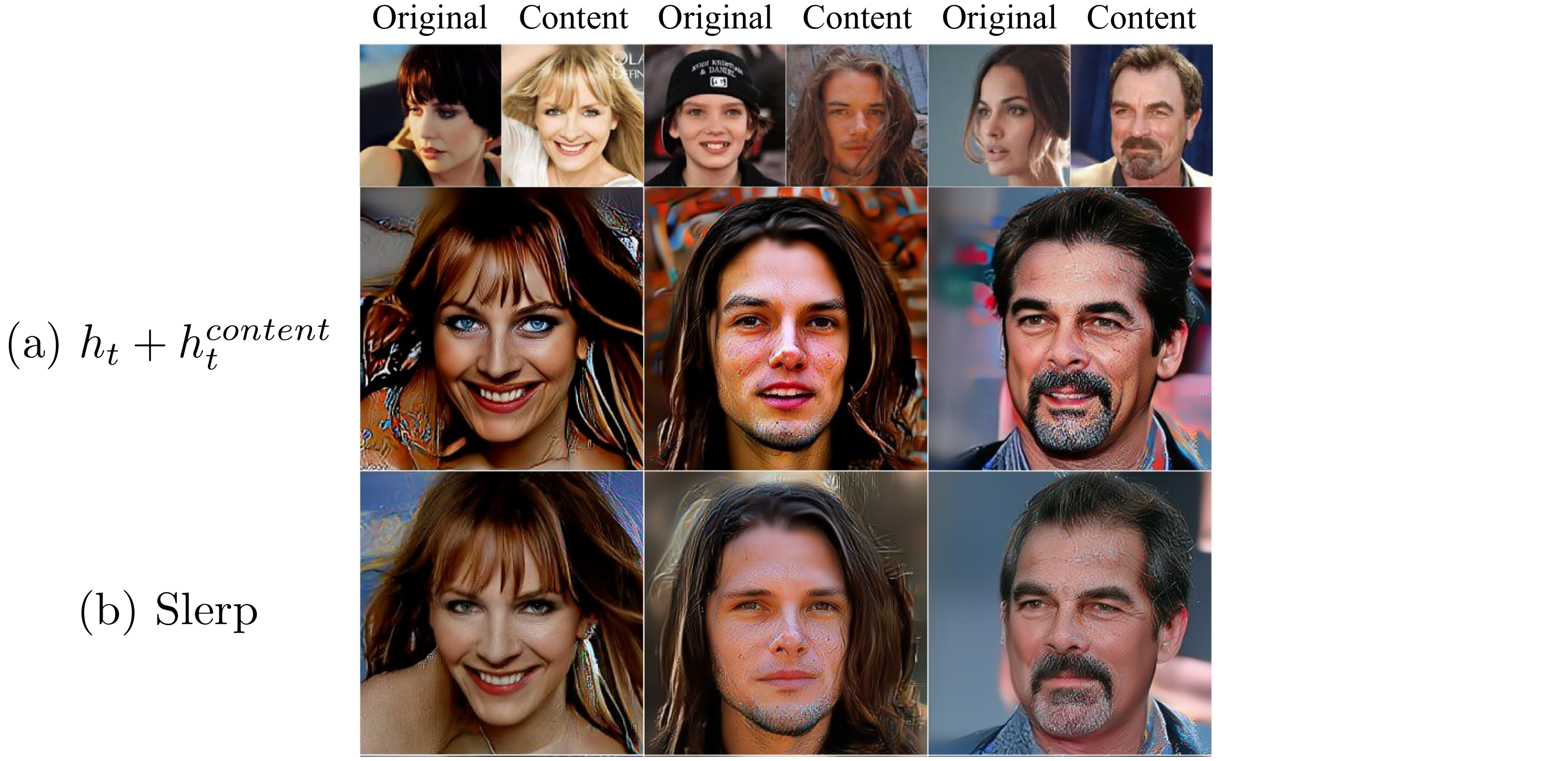}
    \vspace{-1.5em}
    \caption{\textbf{Improvement in quality with Slerp.} (a) shows the result of $\vht+\vhtcon$. It has some artifacts. (b) shows the result of Slerp with $\gamma=0.5$ brings better quality. Techniques described later are not applied here for fair comparison.}  
    \vspace{-1.0em}
    \label{fig:ablation}
\end{figure}

However, the replacement causes severe distortion in the images. We raise another question: how do we prevent the distortion? Note that Asyrp slightly adjusts $\vht$ with a small change $\deltah_t$. On the other hand, replacing $\vht$ as $\vhtcon{}$ completely removes $\vht{}$. Assuming that the maintenance of $\vht{}$ might be the key factor, we try an alternative in-between: adding $\vhtcon{}$ to $\vht{}$; which is illustrated in \fref{fig:slerp_compare}b. We observe far less distortion in \fref{fig:ablation}a.

With these preliminary experiments, we hypothesize that the replacement and the addition drive the disruption of the inherent correlations in the feature map.
The subsequent sections provide grounding analyses and methods to address the problem.

\subsection{Preserving statistics with Slerp}
\label{sec:slerp}

\begin{figure}[!t]
    \centering
    \includegraphics[width=1\linewidth]{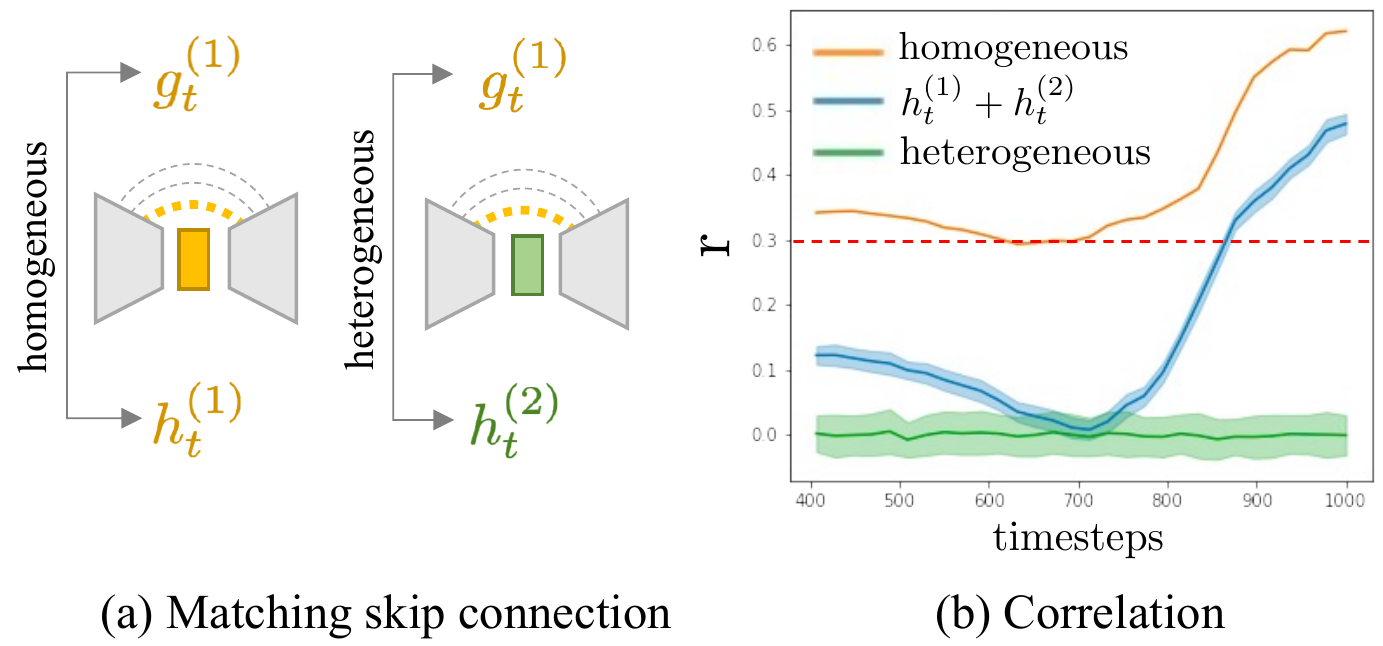}
    \vspace{-1.5em}

    \caption{\textbf{Correlation between $\vht$ and skip connection.} $\vht$ is highly correlated with the matching skip connection. (a) illustrates examples of matching and non-matching skip connections. (b) shows correlation between each $\tildeh_t$ and skip connection. \textbf{r} is Pearson correlation coefficient and p-values of \textbf{r} are less than 1e-15. Non-matching skip connections seriously distort the correlation.}
    \vspace{-1.5em}
    \label{fig:ht_norm}
\end{figure}

In DMs, \thspace{} is concatenated with skip connections and fed into the next layer. However, Asyrp~\cite{kwon2022diffusion} does not take into account the relationship between them.
We observe an interesting relationship between $\vh_t$ and its matching skip connections $\vg_t$ (illustrated in \fref{fig:ht_norm}a) within a generative process and introduce requirements for replacing $\vh_t$. 
We compute two versions of the correlation between the norms, $|\vh_t|$ and $|\vg_t|$:
\begin{equation}
\label{eq:homo}
    r_\text{homo}=\frac{\sum_{i}\left(|\vh^{(i)}|-\bar{|\vh|}\right)\left(|\vg^{(i)}|-\bar{|\vg|}\right)}{(n-1) s_{|\vh|} s_{|\vg|}} 
\end{equation}
\begin{equation}
\label{eq:hetero}
r_\text{hetero}=\frac{\sum_{j\neq i}\left(|\vh^{(j)}|-\bar{|\vh|}\right)\left(|\vg^{(i)}|-\bar{|\vg|}\right)}{(n-1) s_{|\vh|} s_{|\vg|}}
\end{equation}
where $n$ is the number of samples and $s_*$ denotes standard deviation of $*$. We omit $t$ for brevity.

\fref{fig:ht_norm}b shows that $r_\text{homo}$, the correlation between $\vh_t$ and its matching skip connections, is roughly larger than 0.3 and is strongly positive when the timestep is close to $T$. On the other hand, $r_\text{hetero}$, the correlations between $\vh_t$ and the skip connections in different samples, lie around zero.
We try an alternative $\tilde{\vh}=\vh^{(i)}+\vh^{(j)}$ and find its correlation is closer to $r_\text{homo}$ than $r_\text{hetero}$ and it produces less distortion.

Hence, we hypothesize that the correlation between $|\vh|$ and $|\vg|$ should remain consistent after the modification to preserve the quality of the generated images. 
To ensure the correlation of $\tilde{\vh}_t$ equals to $r_\text{homo}$,
we introduce normalized spherical interpolation (Slerp) between $\vht$ and $\vhtcon{}$ : 
\begin{equation}
\label{eq:just_norm}
 \tildeh_t = f(\vht, \vhtcon{}, \gamma) = \text{Slerp}(\vht, \frac{\vhtcon{}}{\left\|\vhtcon{}\right\|}\cdot \left\| \vht \right\|, \gamma),
\end{equation}
where $\gamma\in[0, 1]$ is a coefficient of $\vhtcon{}$. (See \fref{fig:slerp_compare}c.) We note that Slerp requires the inputs to have the same norm. Normalizing $\vhtcon{}$ to match the norm of $\vht$ ensures a consistent correlation between $|\text{Slerp}(\cdot)|$ and $|\vg_t^{(1)}|$ to be the same with the correlation between $|\vht|$ and $|\vg_t^{(1)}|$.
Replacing $\vht$ with $\tilde{\vh}_t$ using Slerp exhibits fewer artifacts and better content preservation, as shown in \fref{fig:ablation}b. Besides the improvement, we can control how much content will be injected by adjusting the $\vht\text{-to-}\vhtcon{}$ ratio through parameter $\gamma_t$ of Slerp.
 We provide an approximation of the total amount of injected content in \sref{supp:cumulative}.

\subsection{Latent calibration}
\label{sec:style_calibration}
So far, we have revealed that mixing features in \thspace{} injects the content.
Although Slerp preserves the correlation between \thspace{} and skip connection, altering only $\vh_t$ with fixed skip connection may arrive at $\tvx_{t-1}$ that could not be reached from $\tvx_t$.
Hence, we propose \textit{\warigari{}} that achieves the similar change due to $\tilde{\vh}_t$ by modifying $\tvx_t$.

Specifically, after we compute $\tvx_{t-1}$, we define a slack variable $\mathbf{v}=\tvx_t+\mathop{d\mathbf{v}}$ and find $\mathop{d\mathbf{v}}$ such that $\Pt(\epsilont(\mathbf{v}))\approx \Pt(\epsilont(\tvx_t|\tildeh_t))$. It ensures $\tvx'_0$ predicted from $\mathbf{v}$ is as similar as possible to $\tvx_0$ predicted from injecting $\tilde{\vh}_t$ to $\tvx_t$.
We model the implicit change from $\tilde\vx_t$ to $\tilde\vx'_t$ that brings similar change by the injection and introduce a hyperparameter $\omega$ that controls the strength of the change. To this end, we define a slack variable $\mathbf{v}=\tvx_t+\mathop{d\mathbf{v}}$ and find $\mathop{d\mathbf{v}}$ such that $\Pt(\epsilont(\mathbf{v}))\approx \Pt(\epsilont(\tvx_t|\tildeh_t))$. With the DDIM equation,
\vspace{-0.5em}
\begin{equation}
\label{eq:Pt}
\sqrt{\alpha_t}\Pt = \tvx_t - \sqrt{1-\alpha_t} \epsilont(\tvx_t),
\vspace{-0.5em}
\end{equation}
we define infinitesimal as 
\vspace{-0.5em}
\begin{equation}
\label{eq:infi}
\sqrt{\alpha_t}\mathop{d\Pt} = \mathop{d\tvx_t} - \sqrt{1-\alpha_t}J(\epsilont)\mathop{d\tvx_t}.
\vspace{-0.5em}
\end{equation}
Further letting $\mathop{d\tvx_t}=\omega\mathop{d\mathbf{v}}$ and
$J(\epsilont)\mathop{d\mathbf{v}}=\mathop{d\epsilont}$ induces  
\begin{equation}
\label{eq:dx}
\mathop{d\tvx_t} = \sqrt{\alpha_t}\mathop{d\Pt} + \omega\sqrt{1-\alpha_t}\mathop{d\epsilont}.  
\end{equation}

Then, we define $\tvx'_t = \tvx_t+d\tvx_t$ and obtain $\tilde\vx'_{t-1}$ by a typical denoising step.

In addition, $\Pt(\epsilont(\tvx'_t))$ in \eref{eq:dx} has larger standard deviation than $\Pt(\epsilont(\tvx_t))$. We regularize it to have the same standard deviation of $\Pt(\epsilont(\tvx_t))$ by 
\begin{equation}
\mathop{d\Pt}= \frac{\Pt' - \bar{\Pt'}}{|\Pt'|}|\Pt|+\bar{\Pt'}-~\Pt(\epsilont(\tilde{\vx}_t)),
\end{equation}
where $\Pt'=\Pt(\epsilont(\tvx'_t))$.
Then we control $\vx'_t$ with an $\omega$. 


When we further expand \eref{eq:dx} by the definition of $\Pt$,
\begin{equation}
\label{eq:dx2}
\mathop{d\tvx_t} \approx (\omega-1)\sqrt{1-\alpha_t}(\epsilont(\tilde{\vx}_t|\tildeh_t)-\epsilont(\tilde{\vx}_t)).
\end{equation}
Interestingly, setting $\omega=1$ reduces $\mathop{d\tvx_t}$ to $\mathbf{0}$, i.e., injection does not occur. And setting $\omega \approx 0$\footnote{$\omega$ can not be 0 because of its definition.} drives $\tvx'_{t-1}$ close to $\tvx_{t-1}$, i.e., \warigari{} does not occur.
Intuitively, by \eref{eq:dx2}, $\tvx'_t$ may share the predicted $\tvx_0$ with $\Pt(\epsilont(\tilde{\vx}_t|\tildeh_t))$ and contains original elements. In other words, we maintain the original elements by adding $\mathop{d\tvx_t}$ directly in \textit{x-space} while the content injection is conducted in \thspace{}. 

\Warigari{} consists of four steps. First, we inject the contents as $\tilde{\vx}_t \to \tilde{\vx}_{t-1}$ with Slerp. Second, we regularize $\Pt$ to preserve the original signal distribution after injection. Third, we solve the DDIM equation $\tvx'_t=\tvx_t+\mathop{d\tvx_t}$ by using \eref{eq:dx}. Finally, we step through a reverse process $\tvx'_t \to \tvx'_{t-1}$. In summary, we obtain target $\tilde{\vx}_{t-1}$ by Slerp and generate $\tvx'_{t-1}$ without feature injection with calculated the corresponding $\tvx'_t$. Please refer to Algorithm \ref{algo:full} for details.

\begin{figure}[!t]
    \centering
    \includegraphics[width=0.8\linewidth]{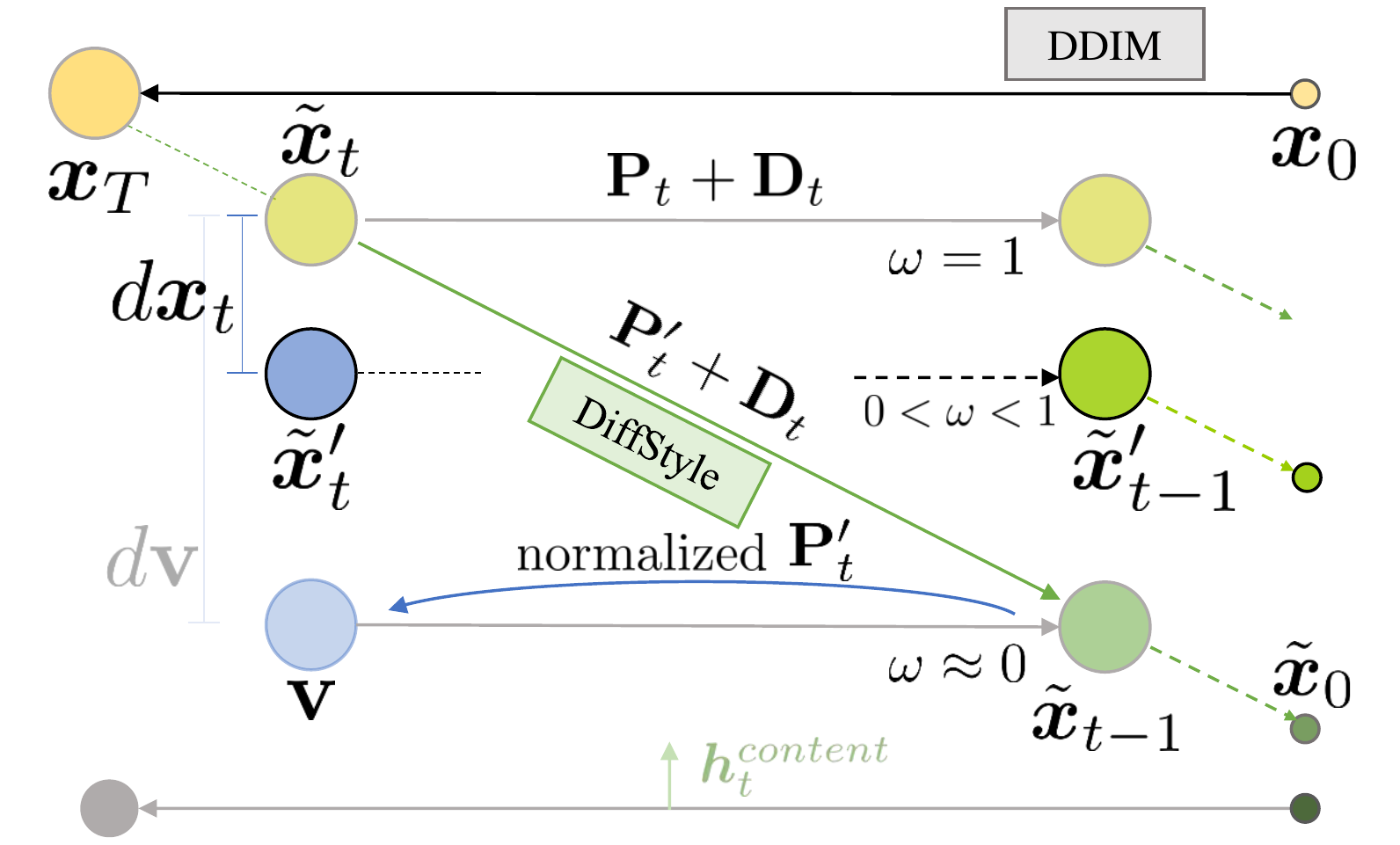}
    \caption{\textbf{\Warigari{}.} The result of DDIM reverse process with given approximated $\tvx'_t$ can be similar to the result of a corresponding injected result $\tvx_{t-1}$. As $\omega$ gets close to 1, more original elements are added through $\mathop{d\vx_t}$. Note that the effect of \warigari{} is different from modifying $\gamma$ because it remains predicted $\tvx_0$ by solving the DDIM equation.}
    \vspace{-1.5em}
    \label{fig:style_calibration}
\end{figure}

\subsection{Full generative process}
We observe that \thspace{} contains content and skip connection from $\vx_T$ conveys the original elements. We utilize this phenomenon for in-domain samples and out-of-domain artistic samples. Note that it is possible to obtain inverted $\vx_T$ from any arbitrary real image. Therefore, even if we use out-of-domain images such as artistic images, \ours{} successfully retain the original elements in the images. Furthermore, local mixing of \thspace{} enables injecting content into the corresponding target area as shown in \fref{fig:limitation_mask}.

 \begin{algorithm}[b!]
    \caption{InjectFusion}
    \label{algo:content_injection}
    \DontPrintSemicolon
    \SetAlgoNoLine
    \SetAlgoVlined
    \SetKwProg{Fn}{Function}{:}{}
    \SetKwFunction{MultiTransfer}{MultiTransfer}
    \KwIn{$\vx_T$ (inverted latent variable from from image $\Istyle{}$),$\{\vhtcon{}\}_{t=t_{edit}}^{T} $(obtained from content image $\Icon{}$), $\epsilon_{\theta}$ (pretrained model), $m$ (feature map mask), $f$ (Slerp)
    } 
    \KwOut{$\tilde{\vx}_0$ (transferred image)}
    \BlankLine
    $\tvx_t \xleftarrow{} \vx_T$
    \For{$t=T,...,1$}{ 
    \If{$t\ge t_{edit}$}{
        Extract feature map $\vht$ from $\epsilon_{\theta}(\tvx{}_t)$;
        $\tildeh_t \xleftarrow{} f((m \otimes \vht), (m \otimes \vhtcon), \gamma)$ \par
        \qquad \qquad \qquad \qquad \qquad 
        $\oplus (1-m) \otimes \vht$\
        $\tilde{\epsilon} \xleftarrow{} \epsilon_{\theta}(\tvx_t | \tildeh_t)$, 
        $\epsilon \xleftarrow{} \epsilon_{\theta}(\tvx_t)$ \\
        Adapt Latent calibration (Algorithm \ref{algo:full})
    }
    \Else{
        $\tilde{\epsilon} = \epsilon \xleftarrow{} \epsilon_{\theta}(\tvx_t)$, 
    }
    $\tvx_{t-1}\xleftarrow{}\sqrt{\alpha_{t-1}} (\frac{\tvx{}_{t}-\sqrt{1-\alpha_{t}}\tilde{\epsilon}}{\sqrt{\alpha_{t}}}) 
    +\sqrt{1-\alpha_{t-1}}\epsilon$
    }
\end{algorithm}

For the local mixing, each $\vht$ is masked before Slerp and the mixed $\vht$ is inserted into the original feature map. We provide Algorithm \ref{algo:content_injection} for them and an illustration of spatial $\vh_t$ mixing in \fref{fig:spatial_slerp}. Note that we omit \warigari{} in the algorithm for simplicity. The full algorithm is provided in Appendix Algorithm~\ref{algo:full}.

\def\NoNumber#1{{\def\alglinenumber##1{}\State #1}\addtocounter{ALG@line}{-1}}

\section{Experiments}
In this section, we present analyses on \ours{} and showcase our applications.
\paragraph{Setting} We use the official pretrained checkpoints of DDPM++ \cite{song2020score,meng2021sdedit} for CelebA-HQ~\cite{karras2018progressive} and LSUN-church/-bedroom~\cite{yu2015lsun}, iDDPM~\cite{nichol2021improved} for AFHQv2-Dog~\cite{choi2020stargan}, and ADM with P2-weighting~\cite{dhariwal2021diffusion,choi2022perception} for \textsc{MetFaces}~\cite{karras2020training} and ImageNet~\cite{imagenet}. The images have a resolution of $256\times256$ pixels. We freeze the model weights.
We use $\tedit$=400, $\omega$=0.3, $\gamma$=0.6, and $\tboost$=200 to produce high-quality images. For more implementation details, please refer to \aref{supp:implement}.

\paragraph{Metrics} GRAM loss (style loss) \cite{gatys2016image} indicates the style difference between the original image and the resulting image. ID computes the cosine similarity between face identity~\cite{deng2019arcface} of the content image and the resulting image to measure content consistency. Fr\'echet Inception Distance (FID)~\cite{heusel2017gans} provides the overall image quality. To compute FID, we compare generated 5K images from fixed 5K original-content image pairs using 50 steps of the reverse process and 25k images from the training set of CelebA-HQ without the overlap of the pairs.

\begin{figure}[!t]
    \centering
    \includegraphics[width=1\linewidth]{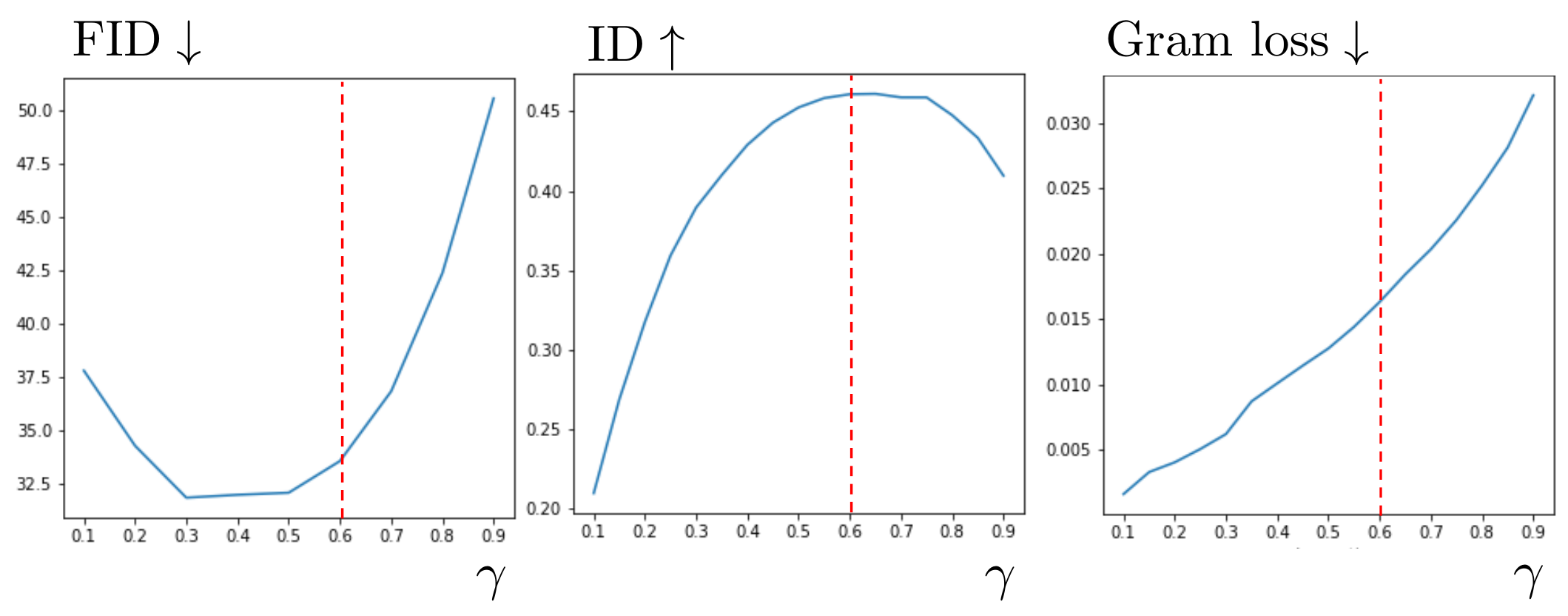}
    \vspace{-2.0em}
    \caption{\textbf{Choice of $\gamma$.} (b) shows that $\gamma$ should be less than $0.6$ since the ID change via content injection converges at the point. If $\gamma > 0.6$, the resulting image only departs from the original image and suffers quality degradation without any advantage.}
    \vspace{-1.0em}
    \label{fig:gamma_id_style}
\end{figure}

\subsection{Analyses}
\label{sec:analyses}
In this section, we define what elements come from the original and the content image. We provide a guideline for choosing the content injection ratio $\gamma$ considering both quality and content consistency. We also show the versatility of latent calibration and propose the best interval for editing. Furthermore, we provide quantitative results that support assumptions suggested in \sref{sec:method}: \thspace{} has content elements.

\begin{table}[t]
\footnotesize
\centering
\resizebox{\columnwidth}{!}{

\begin{tabular}{l|cccccccc}
      [\%] & \multicolumn{1}{l}{Nose} & \multicolumn{1}{l}{Eyes} & \multicolumn{1}{l}{Jaw line} & \multicolumn{1}{l}{Expression} & \multicolumn{1}{l}{Hair color} & \multicolumn{1}{l}{Glasses} & \multicolumn{1}{l}{Skin color} & \multicolumn{1}{l}{Make up} \\ \hline
Original   & 28.06                     & 43.57                    & 24.67                         & 36.73                                  & \textbf{95.74}                 & 5.63                        & \textbf{94.15}                 & \textbf{90.60}              \\
Content & \textbf{71.94}           & \textbf{56.43}           & \textbf{75.33}               & \textbf{63.27}                        & 4.26                           & \textbf{94.37}               & 5.85                           & 9.30                       
\end{tabular}%
}
\resizebox{\columnwidth}{!}{
}
\caption{\textbf{User study to define content} We conduct the user study with 50 participants. Users choose where the attributes of the resulting images come from.}
\vspace{-1em}
\label{tab:clip}
\end{table}

\paragraph{Definition of content}
We measured the CLIP score on CelebA attributes to reveal what information comes from the content and original images. We classify the attribute of the mixed image as closer to the original or content image with the CLIP score. 
In short, content includes \textit{glasses, square jaw, young, bald, big nose, and facial expressions} and the remaining elements include \textit{hairstyle, hair color, bang hair, accessories, beard, and makeup}. Please see the details in \aref{supple:moredetail}.
Furthermore, we conduct a user study in \tref{tab:clip} to support the result of the CLIP score. It aligns with the results using CLIP score for classifying.

We define the retained elements of the original image as the color-dependent attributes and the content as the semantics and shape.
\fref{fig:supple_church} and \fref{fig:supple_bedroom} show that DMs trained on the scenes with complex layouts have different notions of content and retained elements: rough shapes of churches are considered as content and room layouts including the location of beds are considered as contents.

\paragraph{Content injection ratio $\mathcal{\gamma}$}
\label{sec:gamma}
We suggest that the original $\vht$ should be partially kept in \sref{sec:replace}. \fref{fig:gamma_id_style} supports that the content injection ratio $\gamma$ should be less than 0.6 for image quality (FID) and preservation of the original image, and $\gamma>0.6$ does not increase ID similarity.
We provide more observations on $\gamma$ in \aref{supp:gamma}.

\begin{figure}[t]
    \centering
    \includegraphics[width=0.7\linewidth]{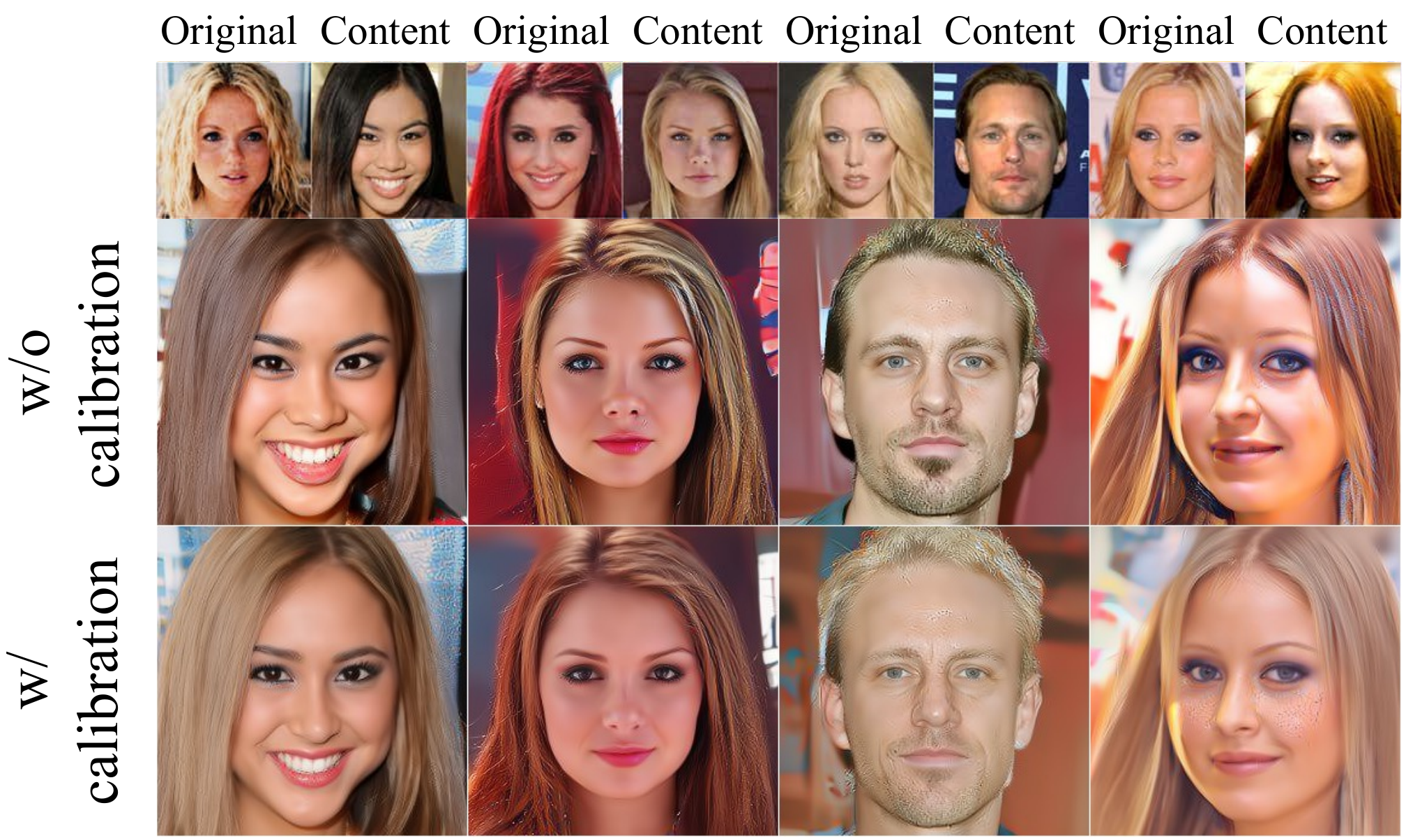}
    \vspace{-0.5em}
    \caption{\textbf{Effectiveness of \Warigari{}.} \Warigari{} recovers elements of original images while preserving content elements. We do not use other techniques such as quality boosting for comparison.}
    \vspace{-1em} 
    \label{fig:warigari_qual}
\end{figure}

\paragraph{The effect of \warigari{}}
\fref{fig:warigari_qual} shows
that \warigari{} leads to a better reflection of the original elements such as makeup and hair color. 
Note that, depending on the latent calibration strength $\omega$, there is a trade-off relationship between Gram loss and ID similarity as well as FID. We report them at various $\omega{}$ in \fref{fig:warigari_quan}.
We discover that increasing $\omega{}$ favors preserving the original images. More details including the efficiency of adapting latent calibration to other methods, Plug-and-Play\cite{tumanyan2023plug} and MasaCtrl\cite{cao2023masactrl}, can be found in \aref{supp:omega}.

\begin{table}[!t]
\centering
\begin{tabular}{l|ccc}
\multicolumn{1}{l|}{} & FID $\downarrow$  & ID $\uparrow$    & Gram loss $\downarrow$ \\ \hline
$\vht{} + \vhtcon{}$                & 49.94 & 0.3581 & 0.0415     \\
Lerp                  & 36.89 & 0.4040 & 0.0318     \\
Slerp                 & \textbf{32.09} & \textbf{0.4390} & \textbf{0.0310}             
\end{tabular}
\caption{\textbf{Performance of various configurations} Slerp improves FID, ID similarity between target content images and synthesized images over other methods. }
\vspace{-0.5em}
\label{tab:ablation}
\end{table}

\begin{figure}[!t]
    \centering
    \includegraphics[width=1.0\linewidth]{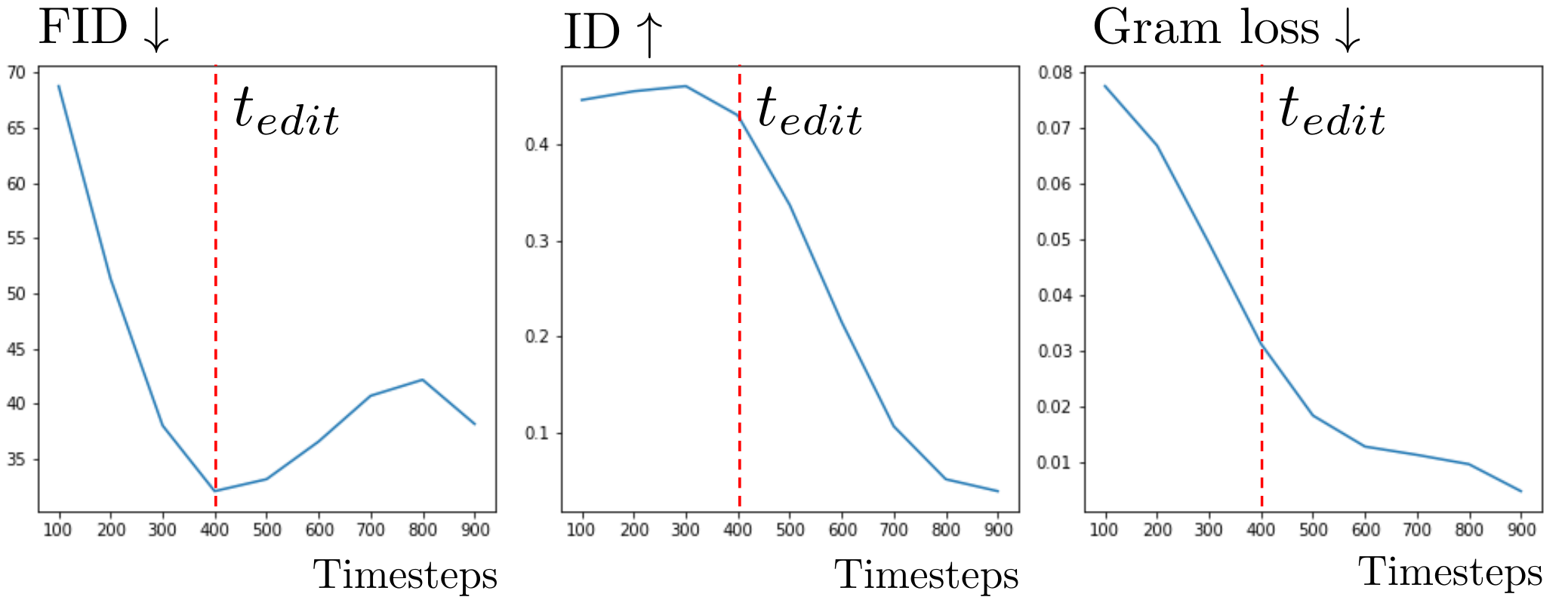}
    \vspace{-1.5em}
    \caption{\textbf{Choice of $\tedit$} We observe that $\tedit=400$ shows the best quality.} 
    \vspace{-0.5em}
    \label{fig:t_end}
\end{figure}  

\begin{figure}[!t]
    \centering
    \includegraphics[width=1.0\linewidth]{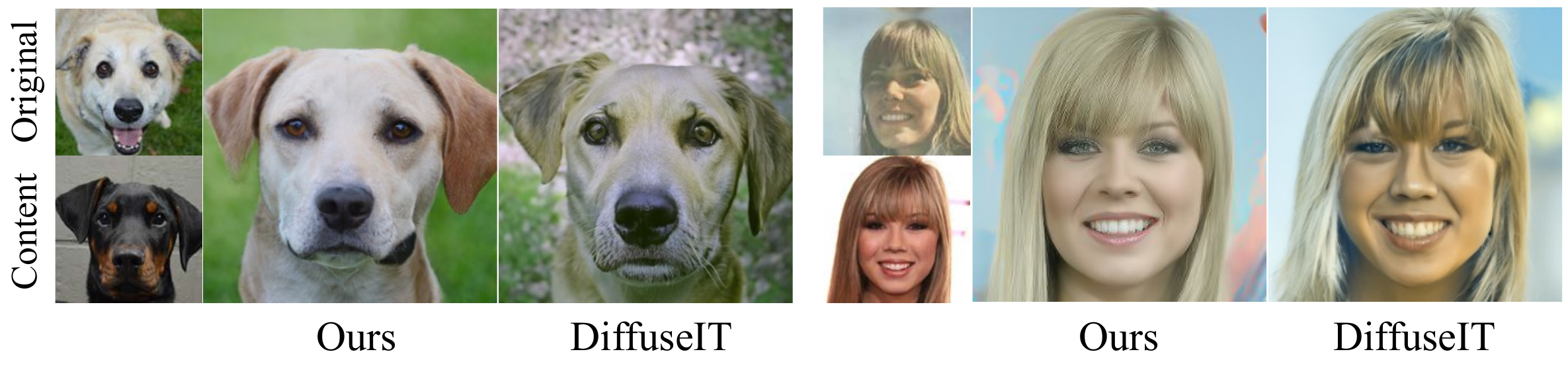}
    \vspace{-1.5em}
    \caption{\textbf{Comparison with DiffuseIT} \ours{} is effective even in situations where there is a large discrepancy between the color distributions of the original image and the content image.} 
    \vspace{-1.5em}
    \label{fig:diffuseIT_compare}
\end{figure}  



    


\paragraph{Quantitative comparison} \tref{tab:ablation} shows the quantitative result of each configuration investigated in \sref{sec:method}. Reconstruction reports FID of the official checkpoint of DDPM++ \cite{meng2021sdedit} through its forward and reverse process without any modification on \thspace{}. We observe that $\vht+\vhtcon$ harms FID with severe distortion.
Slerp outperforms $\vht+\vhtcon$ in all aspects. 

\tref{tab:ablation} further shows the superiority of Slerp over linear interpolation (Lerp). It implies that the normalization for preserving the correlation between $\vht$ and skips $\vg_t$ is important.
Furthermore, \fref{afig:compare_lerp_slerp} shows that Slerp resolves the remaining artifacts that reside in the resulting images by Lerp. 
Comparison between Slerp and Lerp will be further discussed in \sref{supp:lerp}.

\paragraph{Editing interval $[T, \tedit]$}
We observe that there is a trade-off between ID similarity and Gram loss when using a suboptimal $\tedit$ and specific value of $\tedit$ leads to better FID, as shown in \fref{fig:t_end}. 
We choose $\tedit=400$ for its balance among the three factors.
This choice also aligns with that of Asyrp \cite{kwon2022diffusion} for editing toward unseen domains, which requires a large change, such as injecting content.
Notably, we find that $\tedit=400$ is also suitable for achieving content injection into artistic images.

\paragraph{Choice of the content injection layer}
Except for \thspace{}, the other intermediate layers in the U-Net can be candidate feature spaces for content injection. However, \fref{fig:content_injection_on_other_layer}a shows that content injection works well only on \thspace{}, while it produces artifacts and loses injected content on the other feature spaces. Injecting skip connection while content injection does not alleviate the problems as shown in \fref{fig:content_injection_on_other_layer}b.


\subsection{Qualitative results}
\begin{figure*}
    \centering
    \includegraphics[width=0.9\linewidth]
    {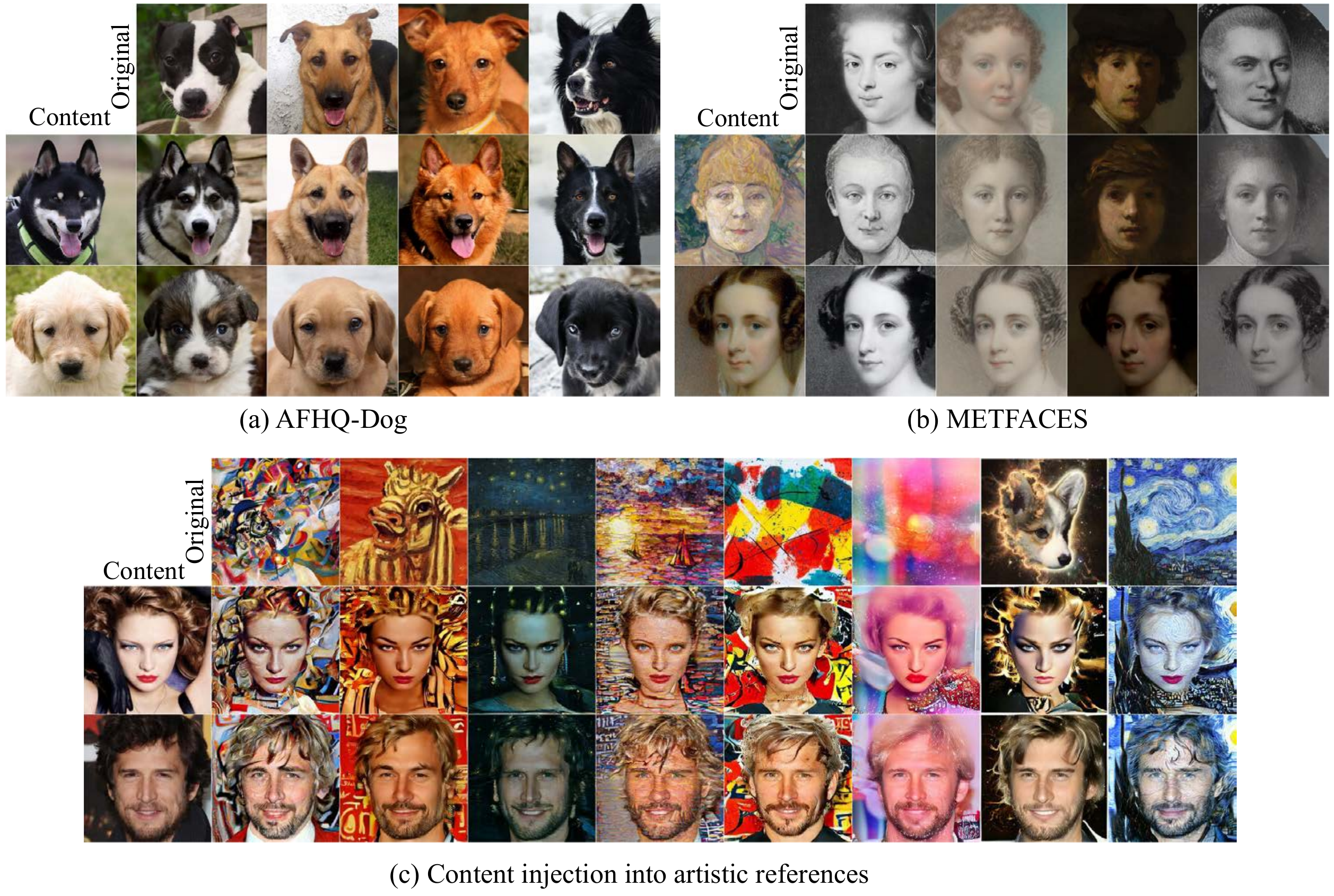}
    \vspace{-0.5em}
    \captionof{figure}{
    \textbf{Qualitative results of \ours{}.} 
        (a), (b) \ours{} allows image mixing by content injection within the trained domain, and (c) out-of-domain artistic references to be original images. All results are produced by frozen pretrained DMs.
 }
    \vspace{-1.0em}
    \label{fig:mixing}
\end{figure*}

\paragraph{In-domain original images}
\fref{fig:mixing}a,b shows \ours{} on AFHQv2-Dog~\cite{choi2020stargan} \textsc{MetFaces}~\cite{karras2020training}.
See \aref{supple:moreresults} for more results on various architectures and datasets.

\paragraph{Artistic original images}
In addition, we can use arbitrary original images, even if they are out-of-domain. \fref{fig:mixing}c shows results with artistic images as style. For the artistic references, we do not use quality boosting \cite{kwon2022diffusion} since they aim to improve the quality and realism of $\vx_0$ which may not be desirable when transferring the elements of an out-of-domain image onto the target image. We provide more results in \aref{supple:moreresults}.

\begin{figure}[!t]
    \centering
    \includegraphics[width=\linewidth]{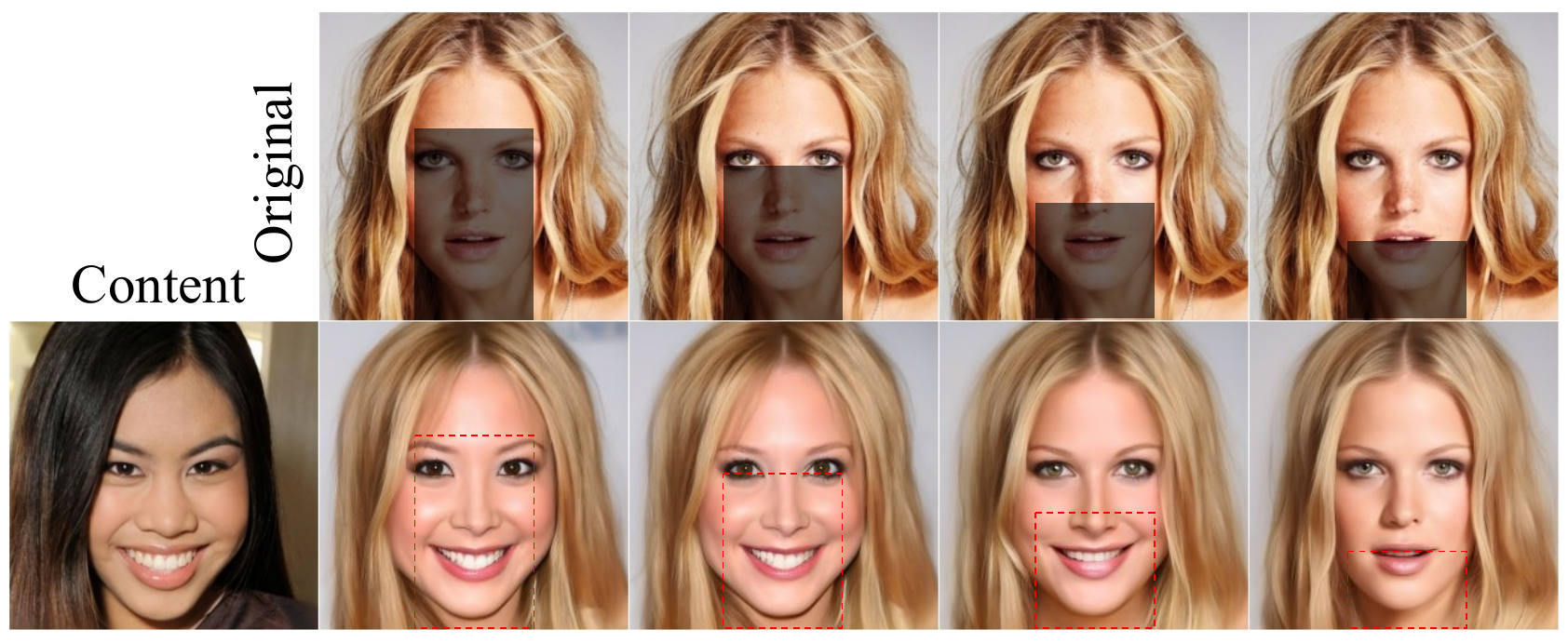}
    \vspace{-1.0em}
    
    \caption{\textbf{Local style mixing with various feature map mask sizes.} Adjusting the size and position of the feature map mask enables to handle the area of content injection, facilitating control of local style mixing.}
    \vspace{-0.5em}
    \label{fig:limitation_mask}
\end{figure}

\subsection{Comparison with existing methods}
 We first note that there is no competitor with perfect compatibility: frozen pretrained diffusion models, and no extra guidance from external off-the-shelf models.
 Still, we compare our content injection with DiffuseIT~\cite{kwon2022diffuseIT} which guides pretrained DMs using DINO ViT \cite{caron2021emerging}.
\fref{fig:diffuseIT_compare} shows that DiffuseIT struggles when there is a large gap between the content image and the original image regarding color distributions. More qualitative comparisons with existing methods~\cite{park2020swapping,kim2021exploiting,deng2021stytr,wu2022ccpl, choi2020stargan, chong2022jojogan} and user study are deferred to \aref{supp:qualitative}.

\begin{figure}[!t]
    \centering
    \includegraphics[width=1\linewidth]{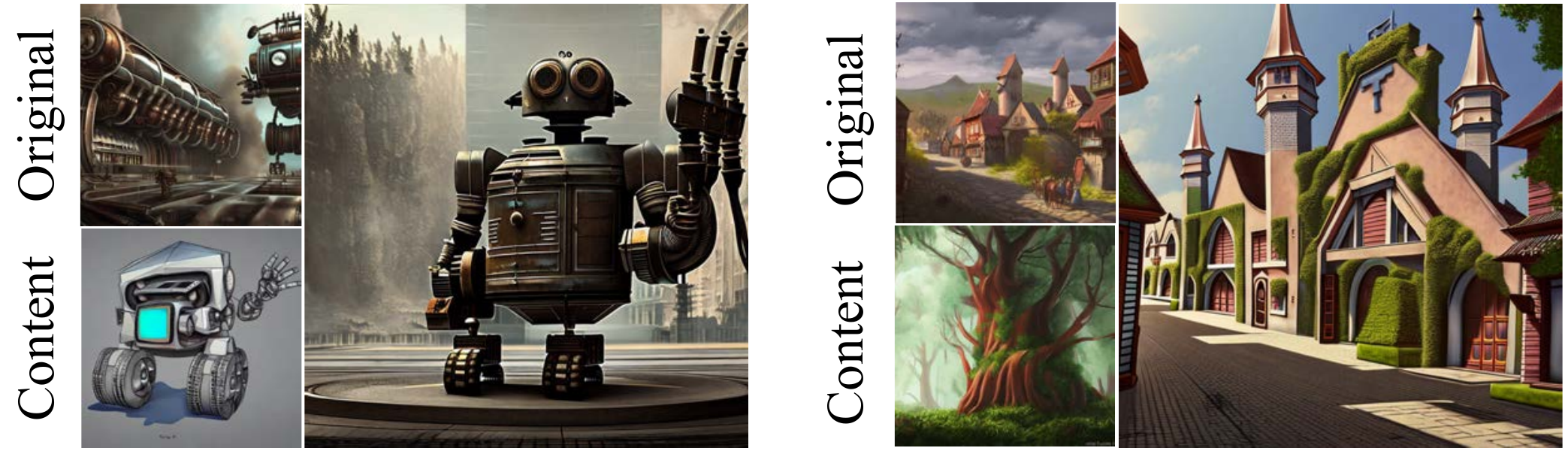}
    \vspace{-1.0em}
    
    \caption{\textbf{\ours{} on Stable diffusion} Although we observe similar phenomenons, the content elements of latent-level DMs is different from pixel-level DMs; More semantic elements is injected to the original image.}
    \vspace{-1.5em}
    \label{fig:stablediffusion}
\end{figure}

\section{Conclusion and discussion}
\label{sec:conclusion}
In this paper, we have proposed a training-free content injection using pretrained DMs. The components in our method are designed to preserve the statistical properties of the original reverse process so that the resulting images are free from artifacts even when the original images are out-of-domain. We hope that our method and its analyses help the research community to harness the nice properties of DMs for various image synthesis tasks.

Although InjectFusion achieves high-quality content injection, the small resolution of the \thspace{} hinders fine control of the injecting region. We provide content injection with various masks in \fref{fig:limitation_mask}.

While out-of-domain images can be used as the original image (i.e., style), injecting content-less out-of-domain images leads to meaningless results. We provide them in \fref{fig:limitation_ood}. We suggest that $\vht$ is not the universal representation for arbitrary content.

In addition, we provide pilot results of \ours{} on Stable diffusion in \fref{fig:stablediffusion}. It works somewhat similarly but the phenomenon is not as clear as in non-latent diffusion models. The bottleneck of Stable diffusion appears to be more semantically rich, possibly due to its diffusion in VAE's latent space.
 Unveiling the mechanisms in latent diffusion models remains our future work. Please refer to \aref{supple:stable} for the details.

Lastly, we briefly discuss the effect of the scheduling strategy of the injecting ratio $\gamma$ in \aref{appendix:gamma_scheduling}. Further investigation would be an interesting research direction.

\subsubsection*{Acknowledgments}
This work was supported by the National Research Foundation of Korea (NRF) funded by the Korean government (MSIT) (RS-2023-00223062).

{\small
\bibliographystyle{ieee_fullname}
\bibliography{egbib}
}

\clearpage{}

\def\supp{1}
    

    

    



\maketitle
\thispagestyle{empty}

\renewcommand{\thetable}{S\arabic{table}}
\renewcommand{\thefigure}{S\arabic{figure}}
\setcounter{figure}{0}
\setcounter{table}{0}

\twocolumn[{
\renewcommand\twocolumn[1][]{#1}
\centering  
\Large
\textbf{Training-free Content Injection using h-space in Diffusion models} \\
\vspace{0.5em}Supplementary Material \\
\vspace{1.0em}
}]
\appendix
\addcontentsline{toc}{section}{Supple}
\renewcommand{\contentsname}{}
\renewcommand{\partname}{} 
\renewcommand{\thepart}{} 
\part{} %

    
 \vspace{-2.0em}

 \begin{algorithm}[h!]
    \caption{\ours{}}
    \label{algo:full}
    \DontPrintSemicolon
    \SetAlgoNoLine
    \SetAlgoVlined
    \SetKwProg{Fn}{Function}{:}{}
    \SetKwFunction{MultiTransfer}{MultiTransfer}
    \KwIn{$\vx_T$ (inverted latent variable from original image $\Istyle{}$),$\{\vhtcon{}\}_{t=t_{edit}}^{T} $(obtained from content image $\Icon{}$), $\epsilon_{\theta}$ (pretrained model), $m$ (feature map mask), $f$ (Slerp), $\omega{}$ (calibration parameter)
    } 
    \KwOut{$\tilde{\vx}_0$ (transferred image)}
    \BlankLine
    $\tvx_t \xleftarrow{} \vx_T$
    \For{$t=T,...,1$}{ 
    \If{$t\ge t_{edit}$}{
    \tcp{step1: Content injection}
        Extract feature map $\vht$ from $\epsilon_{\theta}(\tvx{}_t)$;
        $\tildeh_t \xleftarrow{} f((m \otimes \vht), (m \otimes \vhtcon), \gamma), \omega$ \par
        \qquad \qquad \qquad \qquad \qquad 
        $\oplus (1-m) \otimes \vht$\\
        \tcp{step2: Latent calibration}
        $\tilde{\epsilon} \xleftarrow{} \epsilon_{\theta}(\tvx_t | \tildeh_t)$, 
        $\epsilon \xleftarrow{} \epsilon_{\theta}(\tvx_t)$\\
        $\mu_{\Pt(\tilde{\epsilon})}, \sigma_{\Pt(\tilde{\epsilon})} \xleftarrow{} \Pt(\tilde{\epsilon})$\\
        $\mu_{\Pt({\epsilon})}, \sigma_{\Pt({\epsilon})} \xleftarrow{} \Pt({\epsilon})$\\
        $\Pt' = \mu_{\Pt(\tilde{\epsilon})} + (\Pt(\tilde{\epsilon})-\mu_{\Pt(\tilde{\epsilon})})*\sigma_{\Pt({\epsilon})}$\\
        $d\Pt = \Pt' - \Pt({\epsilon})$\\
        $d\epsilon = \tilde{\epsilon} - \epsilon$\\
        $d\vx = \sqrt{\alpha_t}*d\Pt + \omega*\sqrt{(1-\alpha_t)}*d\epsilon$\\
        $\tilde{\vx_t}'=\tilde{\vx_t}+d\vx$\\
        $\tilde{\epsilon} = \epsilon \xleftarrow{} \epsilon_{\theta}(\tilde{\vx_t}')$
        
    }
    \Else{
        $\tilde{\epsilon} = \epsilon \xleftarrow{} \epsilon_{\theta}(\tvx_t)$, 
    }
    $\tvx_{t-1}\xleftarrow{}\sqrt{\alpha_{t-1}} (\frac{\tvx{}_{t}-\sqrt{1-\alpha_{t}}\tilde{\epsilon}}{\sqrt{\alpha_{t}}}) 
    +\sqrt{1-\alpha_{t-1}}\epsilon$
    }
\end{algorithm}

\section{Implementation details}
\label{supp:implement}
To perform the reverse process for figures, we use 1000 steps, while for tables and plots, we use 50 steps.
During inference, we injecte $\vht{}$ sparsely only at the timesteps where the content injection applied within the 50 inference steps. For the remaining timesteps, we use the original DDIM sampling. This approach enables us to achieve the same amount of content injection across different inference steps.

\begin{figure}[!t]
    \centering
    \includegraphics[width=1\linewidth]{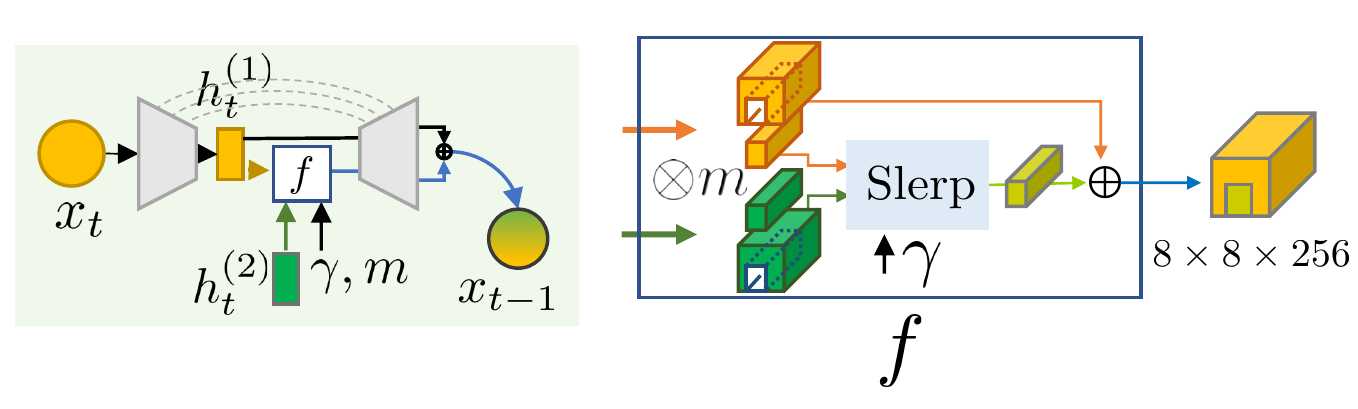}
    \vspace{-0.5em}
    \caption{\textbf{Illustration of local mixing} Mask $m$ determines the area of feature map. Slerp of masked $\vht$ enables content injection into designated space.}
    \vspace{-0.5em}
    \label{fig:spatial_slerp}
\end{figure}

For local mixing, we spatially apply Slerp on $\vht$, which has a dimension of $8 \times 8 \times 256$, as demonstrated in \fref{fig:spatial_slerp}. In face swapping, we use a portion of $\vht{}$ that corresponds to the face area for Slerp. In \sref{sec:method}, we use the editing interval [$T$=1000, $t_{edit}$=400], and do not use quality boosting to eliminate stochasticity for comparison purposes, i.e., $\tboost=0$. 

\section{Varying the strength of content injection}
\label{supp:gamma}

\begin{figure*}
    \centering
    \includegraphics[width=1\linewidth]{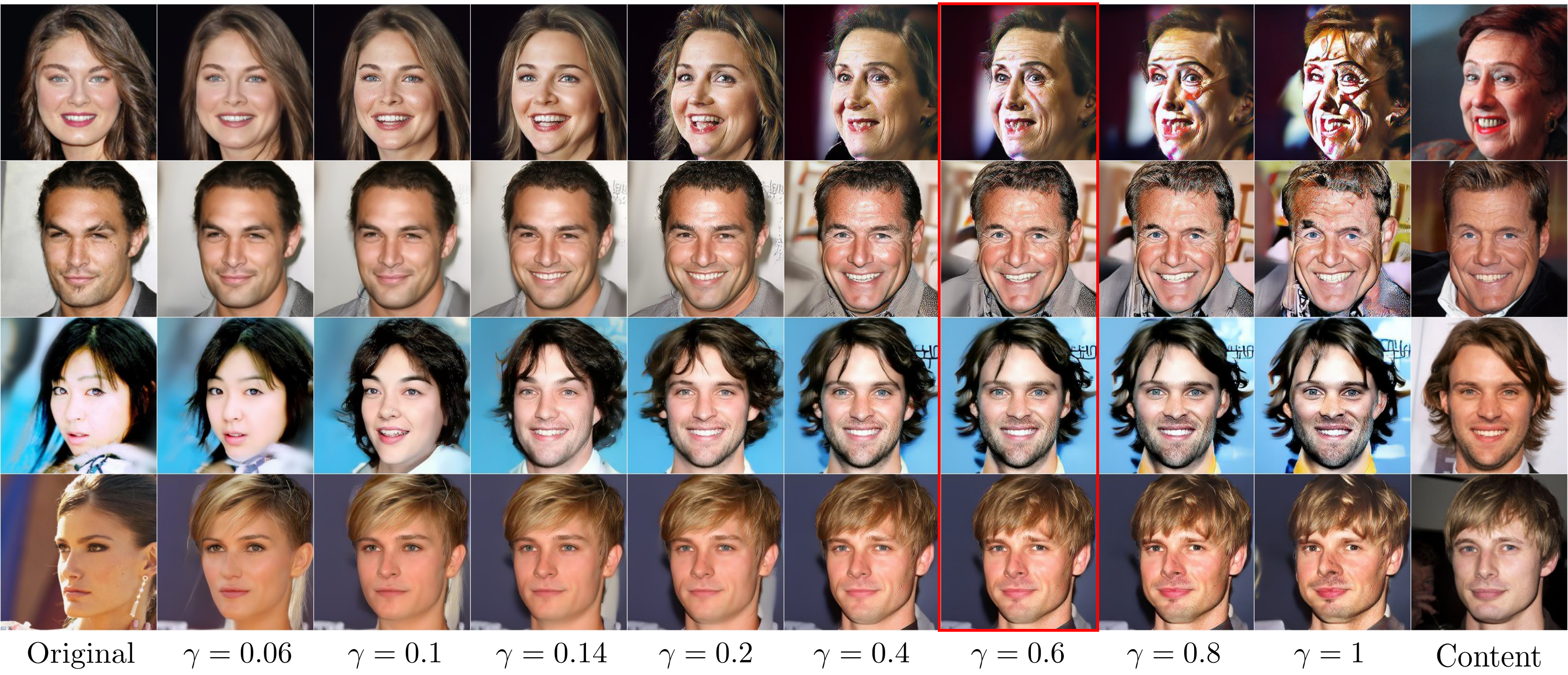}
    \captionof{figure}{$\gamma$ controls how much content will be injected. We do not use other techniques such as quality boosting for comparison.}
    \label{afig:gamma}
\end{figure*}

\fref{afig:gamma} illustrates the results of content injection with different values of Slerp ratio $\gamma{}$. As observed in \fref{fig:gamma_id_style}b, there is a positive correlation between $\gamma{}$ and the amount of content change. However, increasing $\gamma{}>0.6$ barely leads to any content change but degrades the quality of images with distortions and artifacts.
As the recursive injection of content by $\gamma{}$ exponentially decreases the original $\vht{}$ component along the reverse process, according to \eref{eq:cumulative}, we expect linear change of content in the image by linearly controlling $\alpha$ that specifies $\gamma=\alpha^{1/T}$. 

\begin{figure}[t]
    \centering
    \includegraphics[width=1\linewidth]{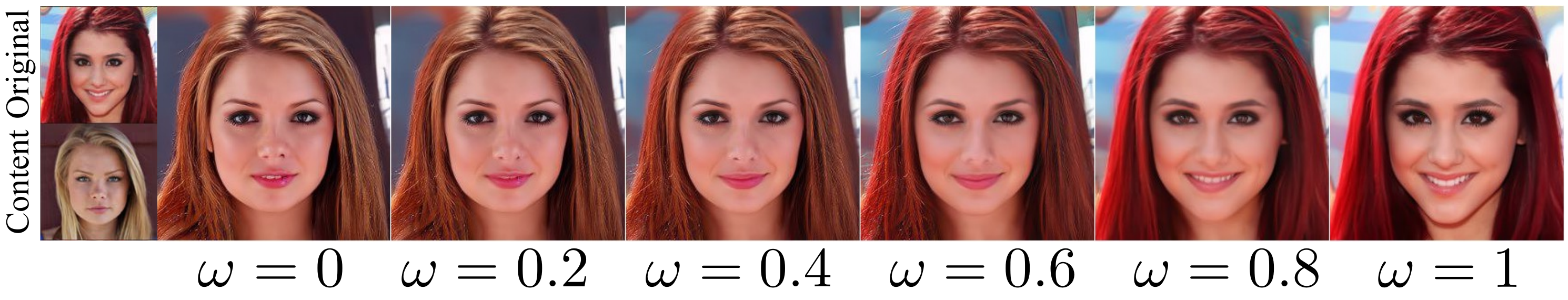}
    \vspace{-0.5em}
    \caption{\textbf{Effect of increasing $\omega{}$}. Increasing $\omega{}$ reflects style elements stronger and $\omega{}=0$ shows the result without latent calibration.}
    \vspace{-0.5em}
    \label{afig:omega_qual}
\end{figure}

\begin{figure}[!t]
    \centering
    \includegraphics[width=1\linewidth]{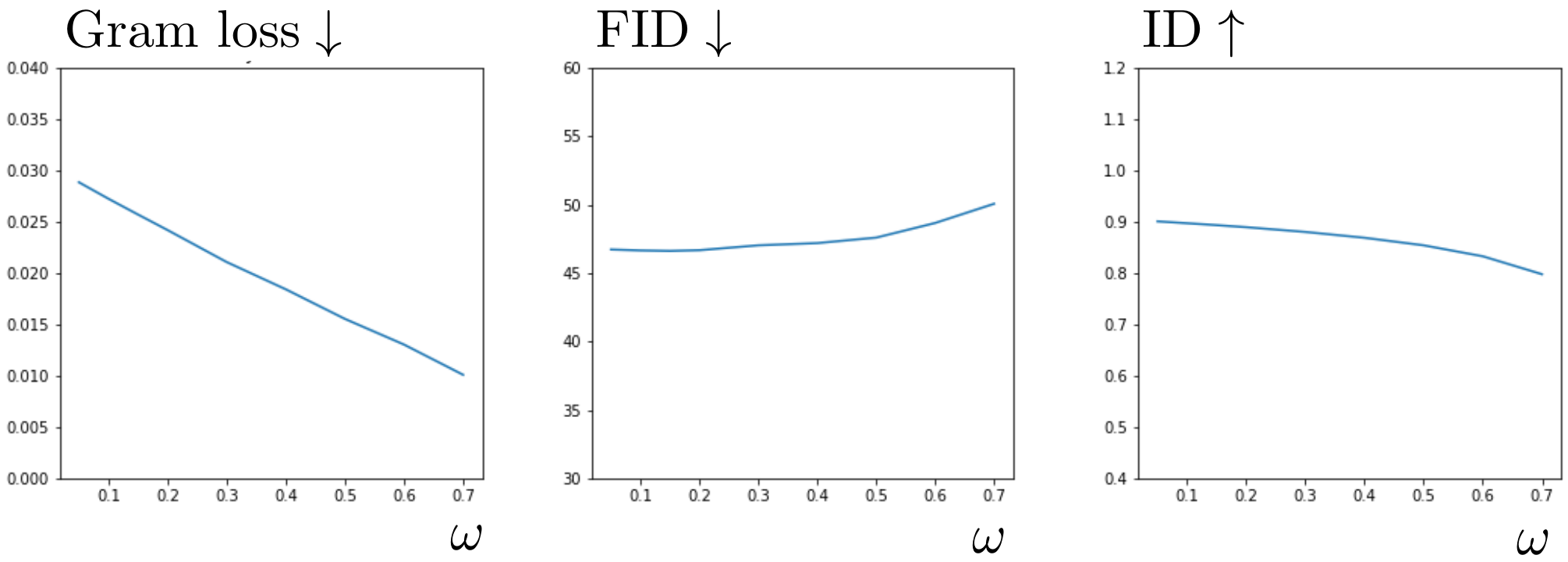}
    \caption{\textbf{Quantitative results of latent calibration with varying $\omega{}$.}
    \Warigari{} ensures that the resulting image remains close to the original image, minimizing content injection loss and preserving image quality.
    }
    \vspace{-1em}
    \label{fig:warigari_quan}
\end{figure}  

\begin{figure}[t]
    \centering
    \includegraphics[width=1\linewidth]{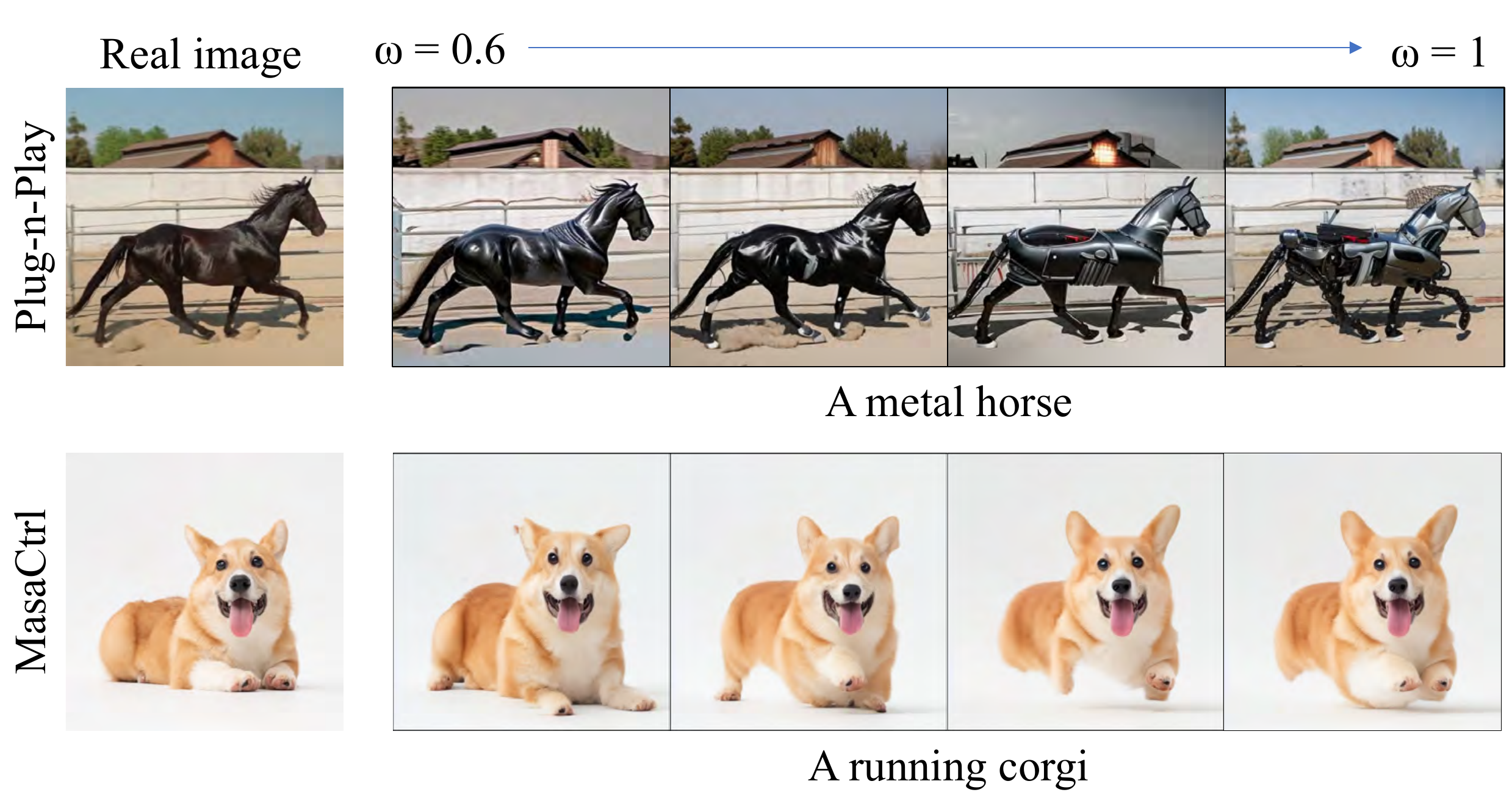}
    \vspace{-0.5em}
    \caption{\textbf{Utilizing latent calibration to other methods.}. Increasing $\omega{}$ reflects injected results stronger when using other methods. For Stable Diffusion, we only use $\omega{}>0.6$. }
    \vspace{-0.5em}
    \label{afig:stylecalibrationothers}
\end{figure}

\section{Effect of latent calibration}
\label{supp:omega}

In this section, we present an analysis of the parameter $\omega{}$ which specifies the strength of the original element. \fref{afig:omega_qual} displays the resulting images with sweeping $\omega{}$. As $\omega{}$ increases, the style elements become more prominent. 
We note that latent calibration with $\omega{}=0$ is not rigorously defined and we report the results without latent calibration when $\omega=0$. In \fref{fig:warigari_quan}, we observe a trade-off between Gram loss and ID similarity, as well as FID, depending on the value of $\omega{}$. However, despite this trade-off, increasing $\omega{}$ results in more effective conservation of the original image.

Because \warigari{} also can control the strength of feature-injected results, we can utilize \warigari{} for other feature-injecting methods, e.g., Plug-and-Play \cite{tumanyan2023plug} and MasaCtrl \cite{cao2023masactrl}. \fref{afig:stylecalibrationothers} shows that increasing $\omega$ increases the strength of editing.

\section{More results and comparison}

\subsection{More qualitative results}
\label{supple:moreresults}

\begin{figure}[t]
    \centering
    \includegraphics[width=0.7\linewidth]{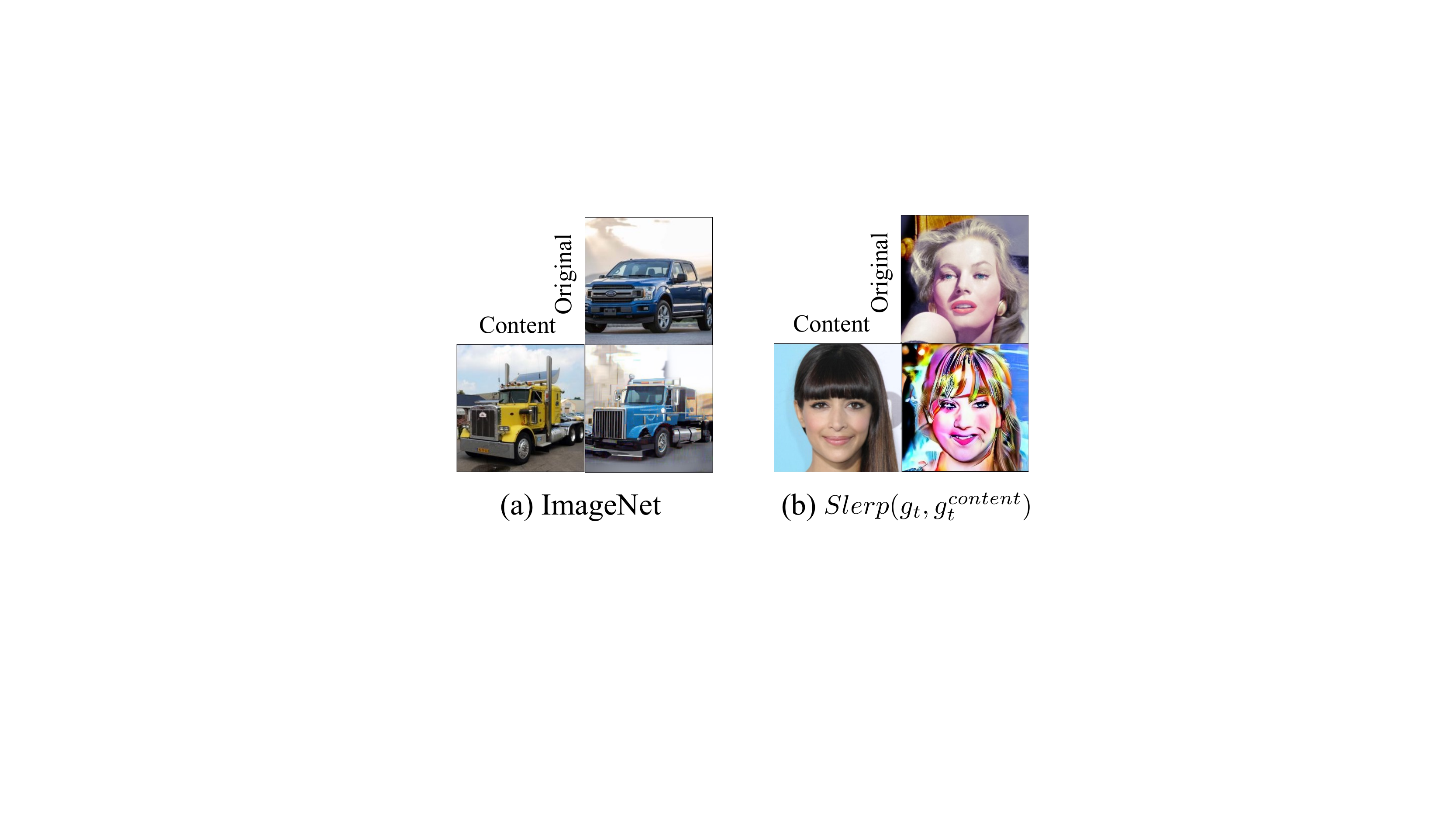}
    \vspace{-1.0em}
    \caption{(a) \ours{} works on ImageNet. (b) Skip connection injection does not provide meaningful results. }
    \label{fig:supple_imagenet_skip_slerp}
\end{figure}

\begin{figure*}
    \centering
    \includegraphics[width=\linewidth]{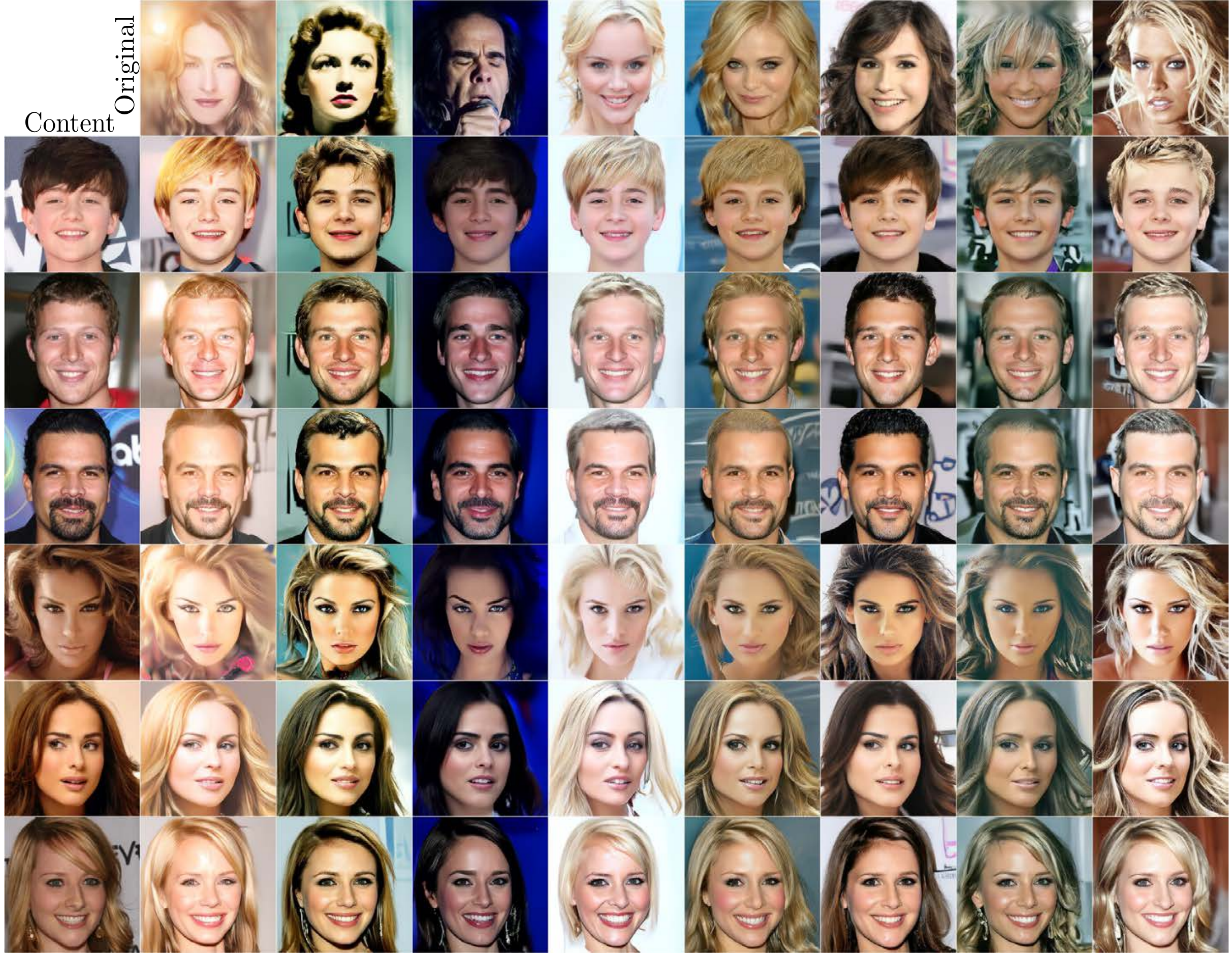}
    \captionof{figure}{Qualitative results of content injection on CelebA-HQ.}
    \label{fig:supple_celeba}
    \vspace{-1em}
\end{figure*}

\begin{figure*}
    \centering
    \includegraphics[width=\linewidth]{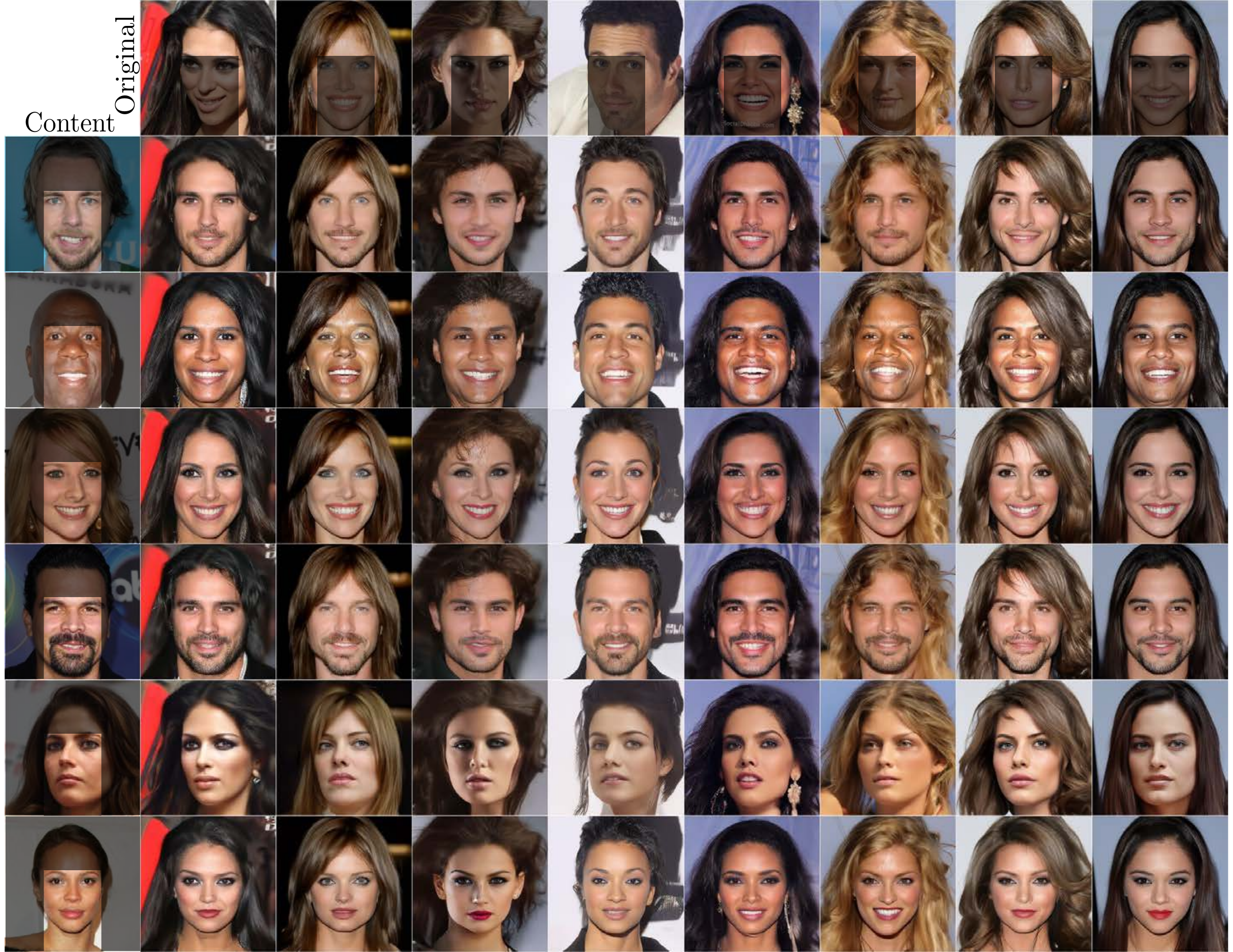}
    \captionof{figure}{Qualitative results of local editing on CelebA-HQ.}
    \label{fig:supple_celeba_mask}
    \vspace{-1em}
\end{figure*}

\begin{figure*}
    \centering
    \includegraphics[width=\linewidth]{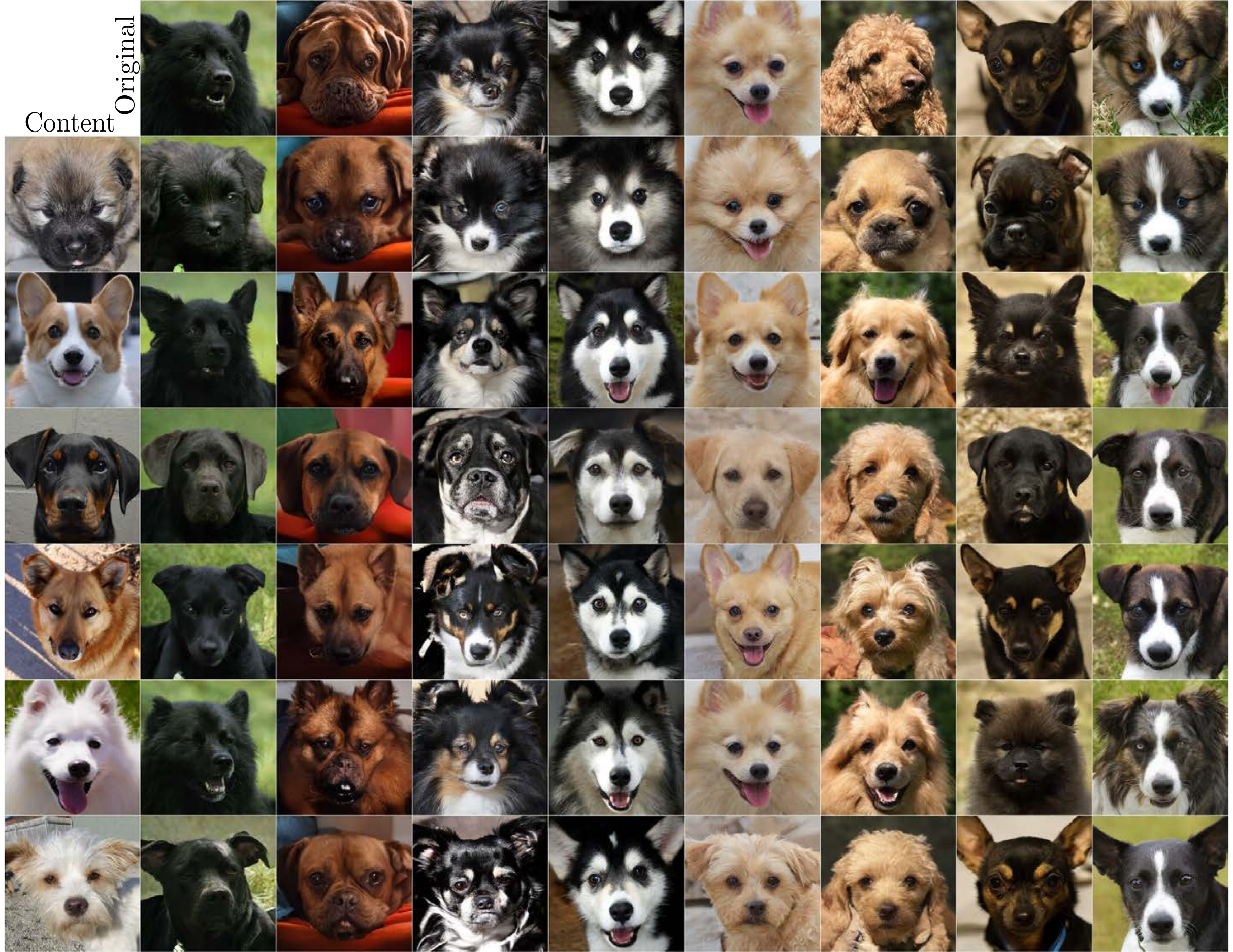}
    \captionof{figure}{Qualitative results of content injection on AFHQ. }
    \label{fig:supple_afhq}
    \vspace{-1em}
\end{figure*}

\begin{figure*}
    \centering
    \includegraphics[width=\linewidth]{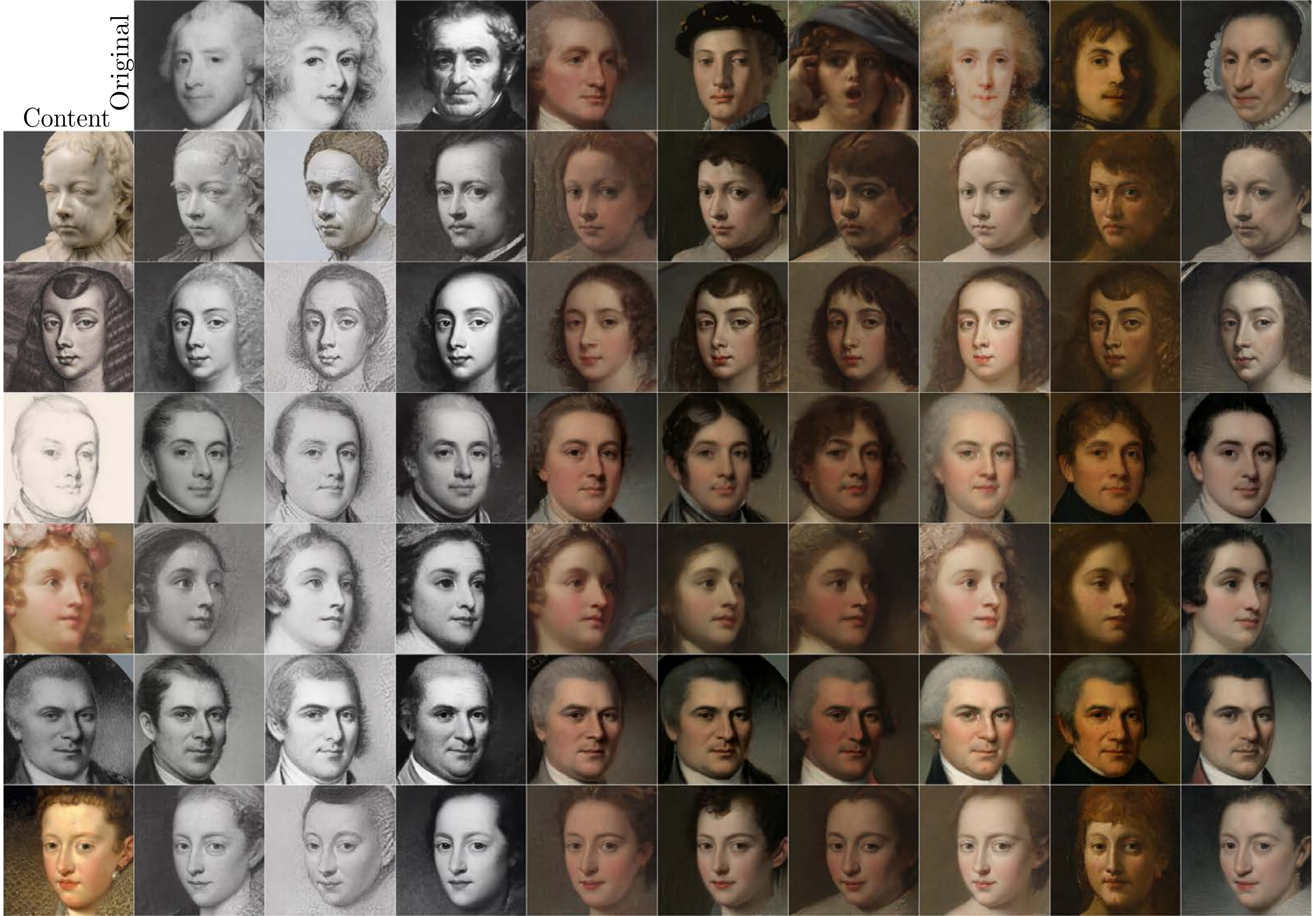}
    \captionof{figure}{Qualitative results of content injection on \metfaces{}.}
    \label{fig:supple_metface}
    \vspace{-1em}
\end{figure*}

\begin{figure*}
    \centering
    \includegraphics[width=\linewidth]{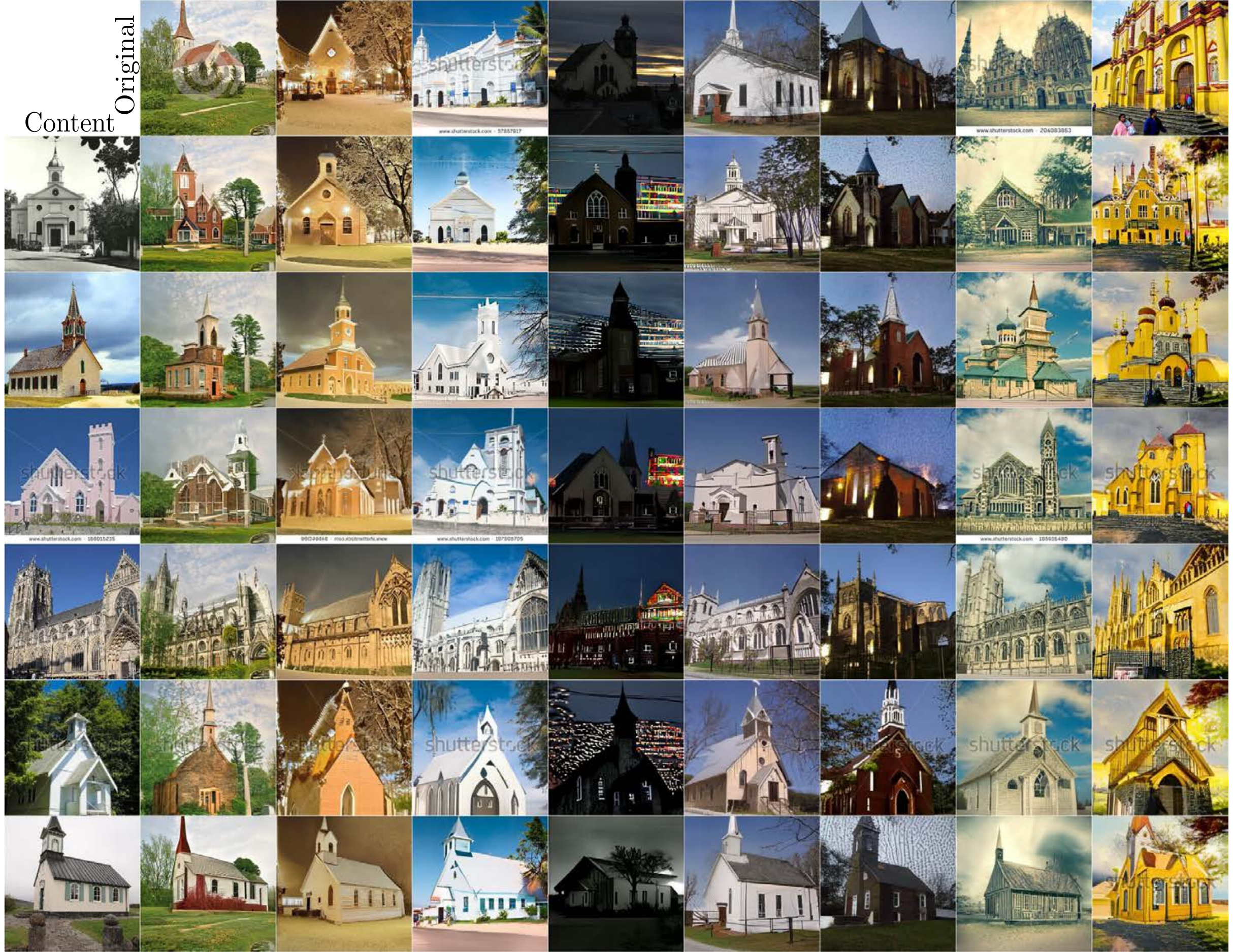}
    \captionof{figure}{Qualitative results of content injection on LSUN-church.}
    \label{fig:supple_church}
    \vspace{-1em}
\end{figure*}

\begin{figure*}
    \centering
    \includegraphics[width=\linewidth]{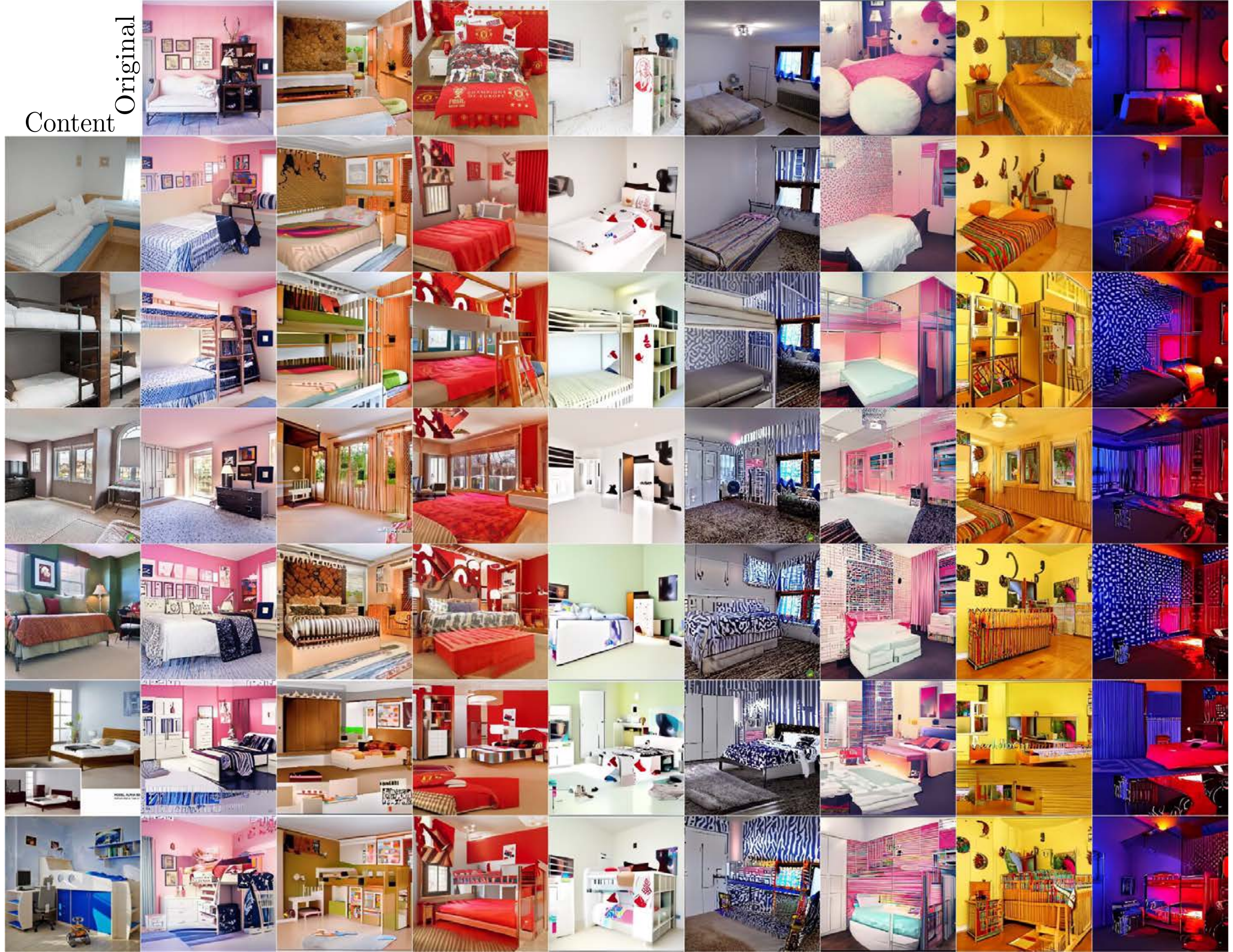}
    \captionof{figure}{Qualitative results of content injection on LSUN-bedroom.}
    \label{fig:supple_bedroom}
    \vspace{-1em}
\end{figure*}

\begin{figure*}
    \centering
    \includegraphics[width=\linewidth]{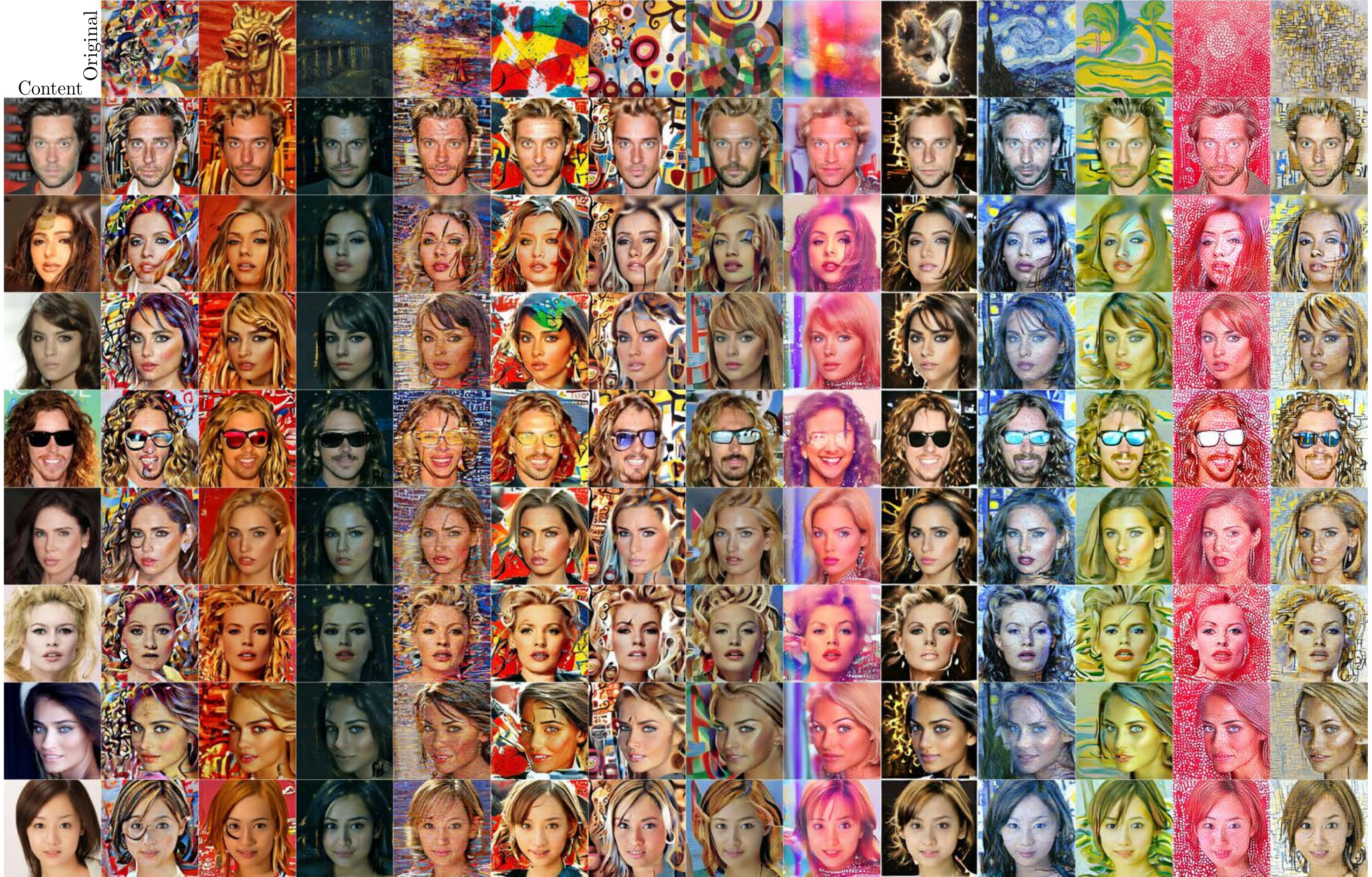}
    \captionof{figure}{Qualitative results of content injection into artistic references with CelebA-HQ .}
    \label{fig:supple_style_transfer_more}
    \vspace{-1em}
\end{figure*}

We provide more qualitative results of CelebA-HQ, AFHQ, \metfaces{}, LSUN-church, and LSUN-bedroom in \fref{fig:supple_celeba}-\ref{fig:supple_style_transfer_more}  (located at the end for compact arrangement).
We also provide a result of ImageNet in \fref{fig:supple_imagenet_skip_slerp}a.
\begin{table}[b]
\centering
\resizebox{\columnwidth}{!}{
\begin{tabular}{l|l|c}
                        & Method               & Preference (\%) \\
                        \hline
\multirow{2}{*}{Content injection}       & Swapping Autoencoder~\cite{park2020swapping} & 40.11      \\
                        & Ours                 & \textbf{59.89}      \\
                        \hline
\multirow{2}{*}{Local content injection} & StyleMapGAN~\cite{kim2021exploiting}          & 33.56      \\
                        & Ours                 & \textbf{66.44}      \\
                        \hline
\multirow{3}{*}{Artistic style transfer} & StyTr$^2$~\cite{deng2021stytr}            & 20.89      \\
                        & CCPL~\cite{wu2022ccpl}                 & 21.44      \\
                        & Ours                 & \textbf{57.67}     \\
\end{tabular}
}
\caption{User study with 90 participants.}
\label{atab:comparison}
\end{table}

\begin{figure*}[t]
    \centering
    \includegraphics[width=\linewidth]{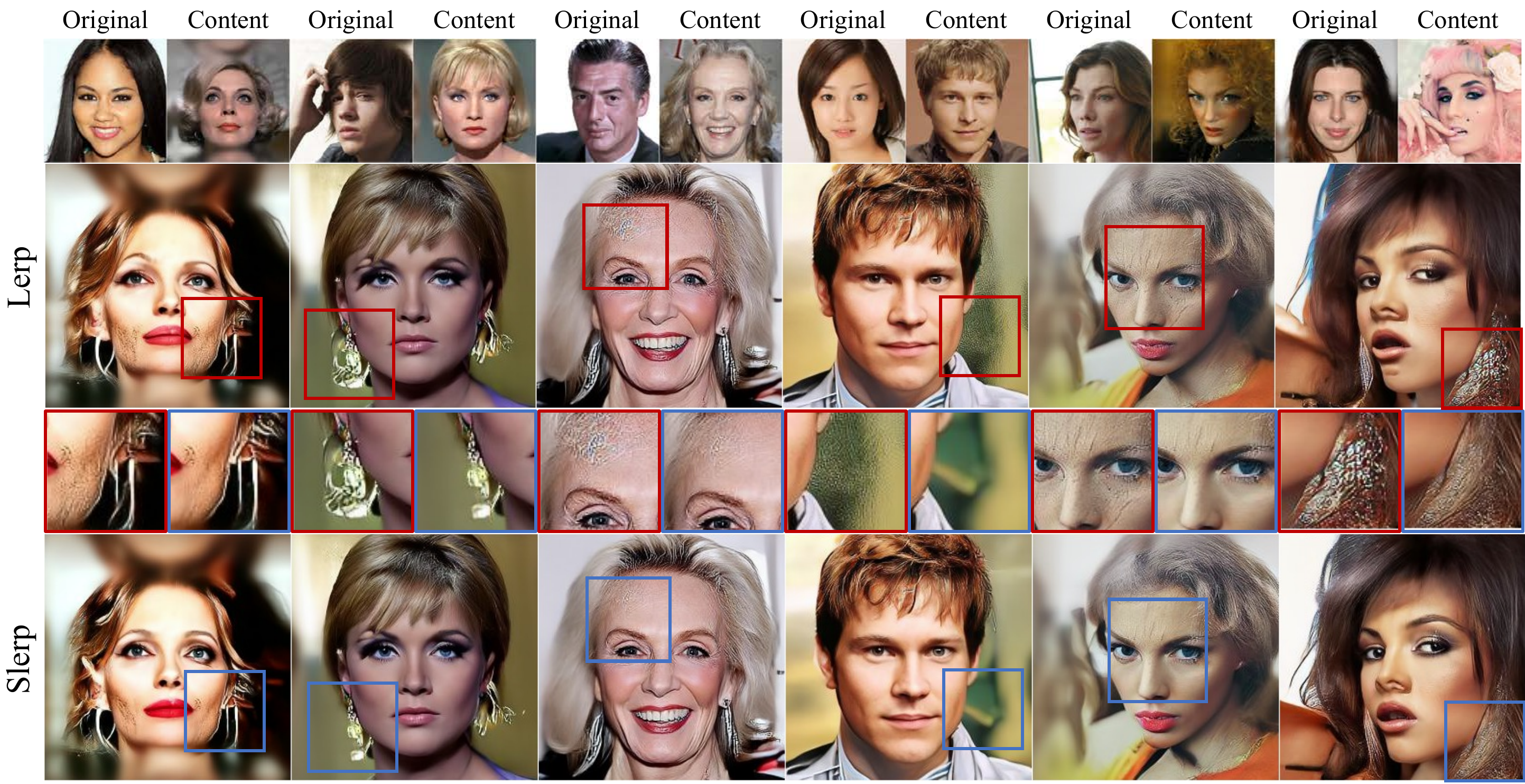}
    \captionof{figure}{\textbf{Comparison between Slerp and Lerp.} Slerp reduces artifacts and distortions in Lerp. Note that We do not use other techniques such as quality boosting to evaluate the effect of Slerp only.}
    \label{afig:compare_lerp_slerp}
\end{figure*}

\subsection{Comparison with the other methods.}
\label{supp:qualitative}

\begin{figure*}
    \centering
    \includegraphics[width=\linewidth]{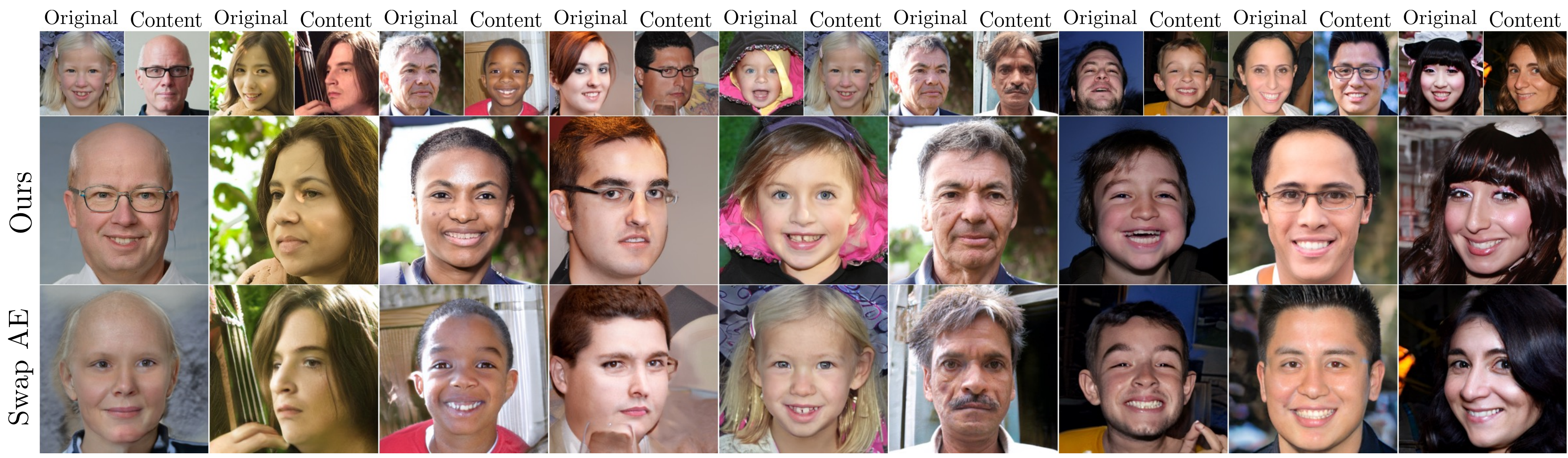}
    \captionof{figure}{\textbf{Qualitative comparison of content injection on FFHQ.} \ours{} is shown to be effective in reflecting content elements while preserving the overall color distribution of the original image. }
    \label{fig:supple_swapping_autoencoder}
    \vspace{-1em}
\end{figure*}

\begin{figure*}
    \centering
    \includegraphics[width=\linewidth]{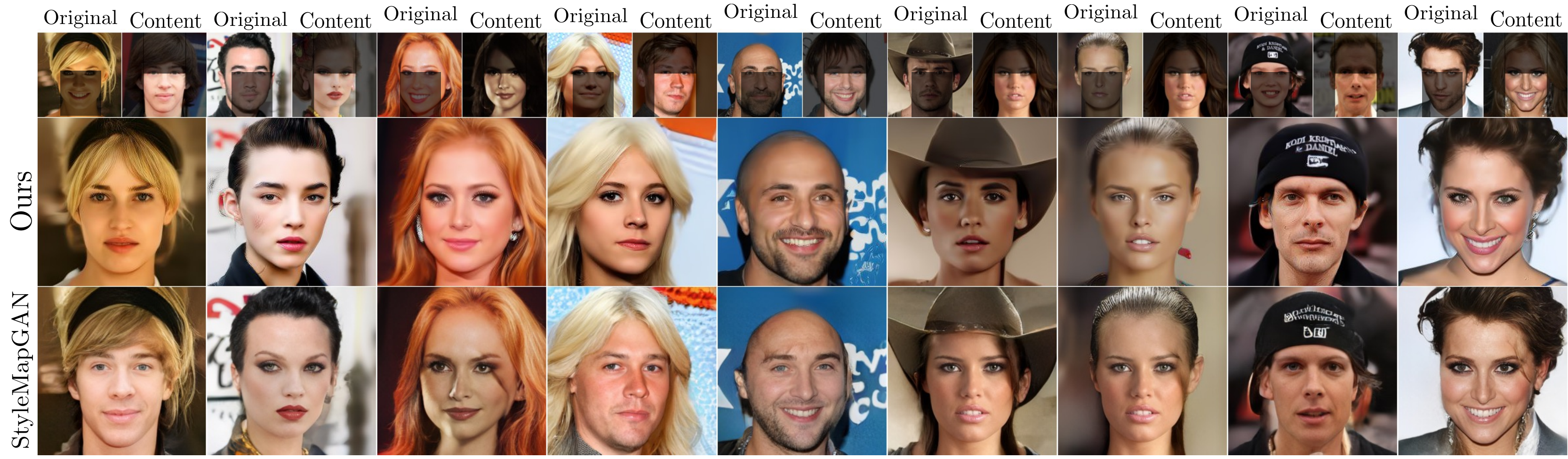}
    \captionof{figure}{\textbf{Qualitative comparison of local mixing on CelebA-HQ.} Despite providing StyleMapGan with detailed segmentation guidance, there are noticeable artifacts in the resulting images, especially at the border lines of the mask. Furthermore, due to the differences in pose between the content and the original images, StyleMapGan struggles to seamlessly integrate the two images, resulting in less-than-optimal outcomes.}
    \label{fig:supple_stylemapgan}
    \vspace{-1em}
\end{figure*}

\begin{figure*}
    \centering
    \includegraphics[width=\linewidth]{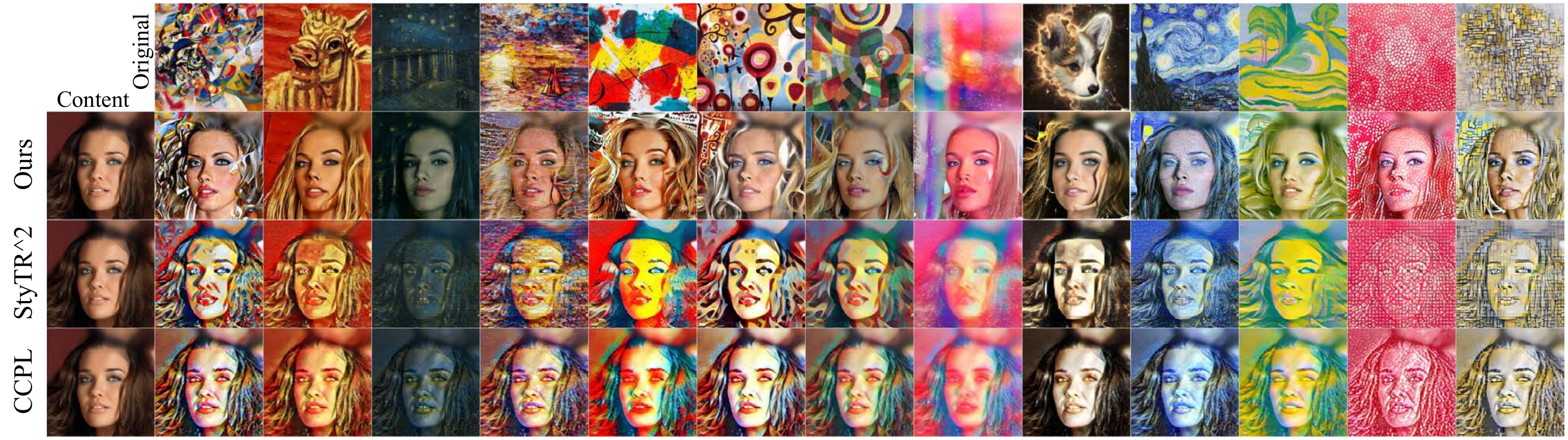}
    \captionof{figure}{\textbf{Qualitative comparison between \ours and style transfer methods with artistic references on CelebA-HQ.} \ours{} allows using images from unseen domains as the original images, enabling the target content can be reflected on the artistic references. \ours{} produces a harmonization-like effect without severe content distortion. Some high-level semantic color patterns of the original images are better reflected by \ours{}  than the others.}
    \label{fig:supple_styletransfer_comparison}
    \vspace{-1em}
\end{figure*}

\tref{atab:comparison} presents the results of a user study conducted with 90 participants to compare our method with existing methods. The participants were asked a question: ``Which image is more natural while faithfully reflecting the original image and the content image?". We randomly selected ten images for content injections and thirty images for style transfer without any curation. The example images are shown in \fref{fig:supple_swapping_autoencoder}-\ref{fig:supple_styletransfer_comparison} (located at the end for clear spacing). Even though \ours{} works on pretrained diffusion models without further training for the task, our method outperforms the others. We selects the recent methods from the respective tasks for comparison. 

\begin{figure}[t]
    \centering
    \includegraphics[width=\linewidth]{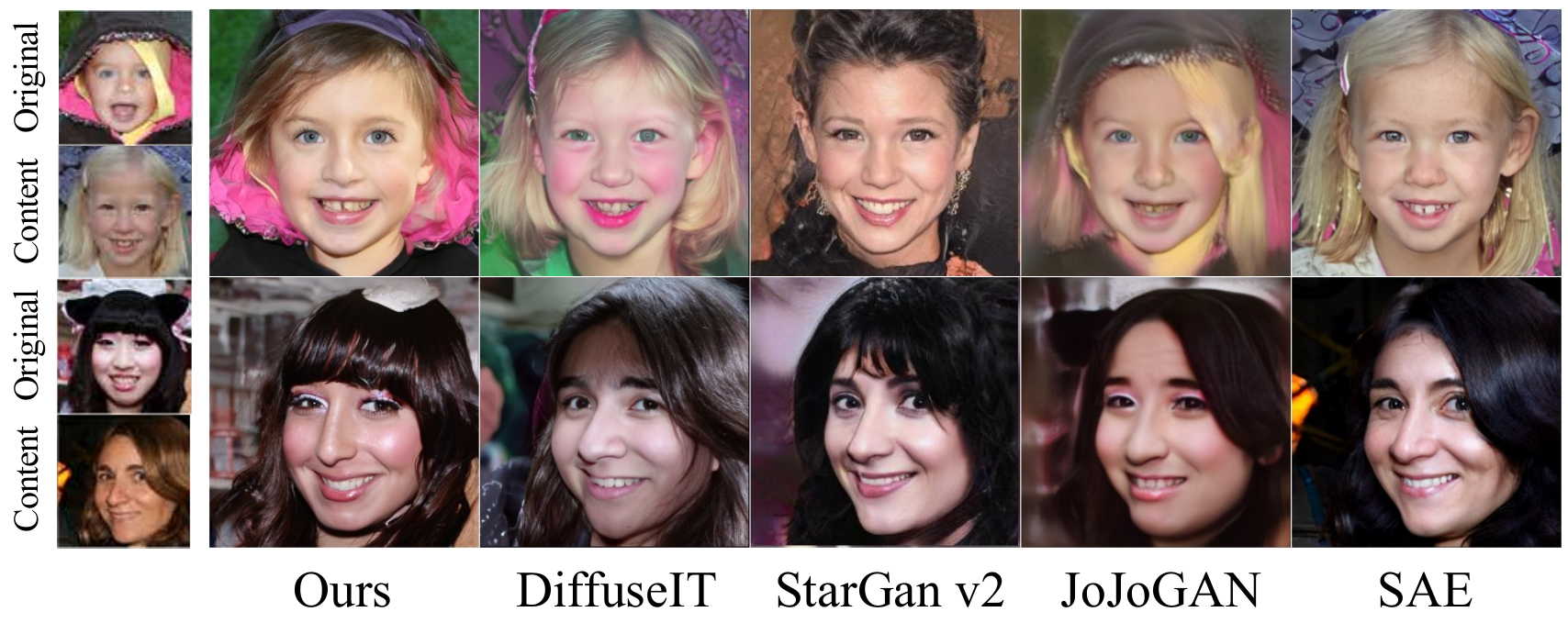}
    \vspace{-1.0em}
    \caption{\textbf{More comparisons} \ours{} shows different mixing strategy compared to the other methods.}
    \label{fig:supple_mixing_compare}
\end{figure}

Although content injection does not define domains of images, it resembles image-to-image translation in that both of their results preserve content of input images while adding different elements.
Therefore, we show the differences between \ours{} and those works in \fref{fig:supple_mixing_compare}. The resulting image of \ours{} well reflects overall color distribution, color-related attributes (e.g. makeup), and non-facial elements (e.g. long hair, bang hair, decorations on a head) of the original images. Ours also reflect facial expression, jawline, and overall pose of the content image. On the other hand, the other works do not accurately reflect color-related attributes from the original images and also ignore fine-grained detail or spatial structure of the original image. They focus on preserving the structure of the content image.

\subsection{Comparison with DiffuseIT}
\begin{figure*}
    \centering
    \includegraphics[width=0.7\linewidth]{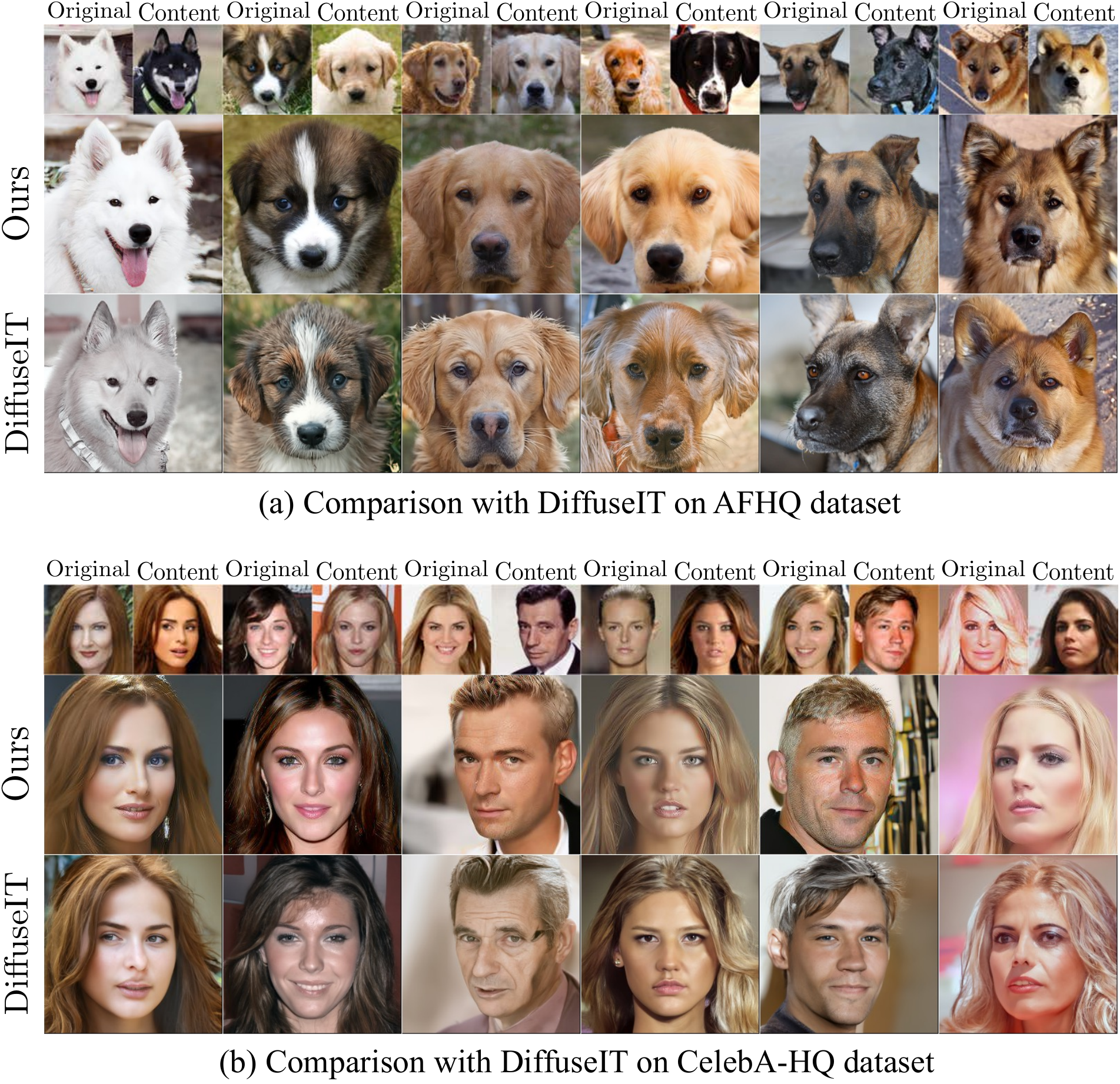}
    \captionof{figure}{\textbf{More qualitative comparison with DiffuseIT.} \ours{} excels in fully and naturally reflecting the original color without creating artificial contrast, particularly when there is a significant gap between the content color and the style color (e.g., black and white). In contrast, DiffuseIT may not fully capture the original color in such cases.}
    \label{fig:supple_diffuseIT}
    \vspace{-1em}
\end{figure*}

We provide more qualitative comparison with DiffuseIT \cite{kwon2022diffuseIT} which uses DINO ViT \cite{caron2021emerging}.
As shown in \fref{fig:supple_diffuseIT}, \ours{} shows comparable results without extra supervision. \ours{} is highly proficient at accurately and authentically reflecting the color of the original image while avoiding artificial contrast, especially when there is a significant difference in color between the content and the original image (e.g., black and white). In contrast, DiffuseIT may not be able to fully capture the color of the original image in these scenarios. This discrepancy is due to the starting point of the reverse process. DiffuseIT utilizes the inverted $\xT$ of the content image to sample and manipulate noise to match the target original image. The large gap in color distribution between the content and original images makes it challenging for DiffuseIT to overcome this difference entirely. Conversely, \ours{} initially samples from the inverted $\xT$ of the original image, making it easier to maintain the color of the original image. The original image is preserved through the skip connection.
 

\begin{figure}
    \centering
    \includegraphics[width=0.8\linewidth]{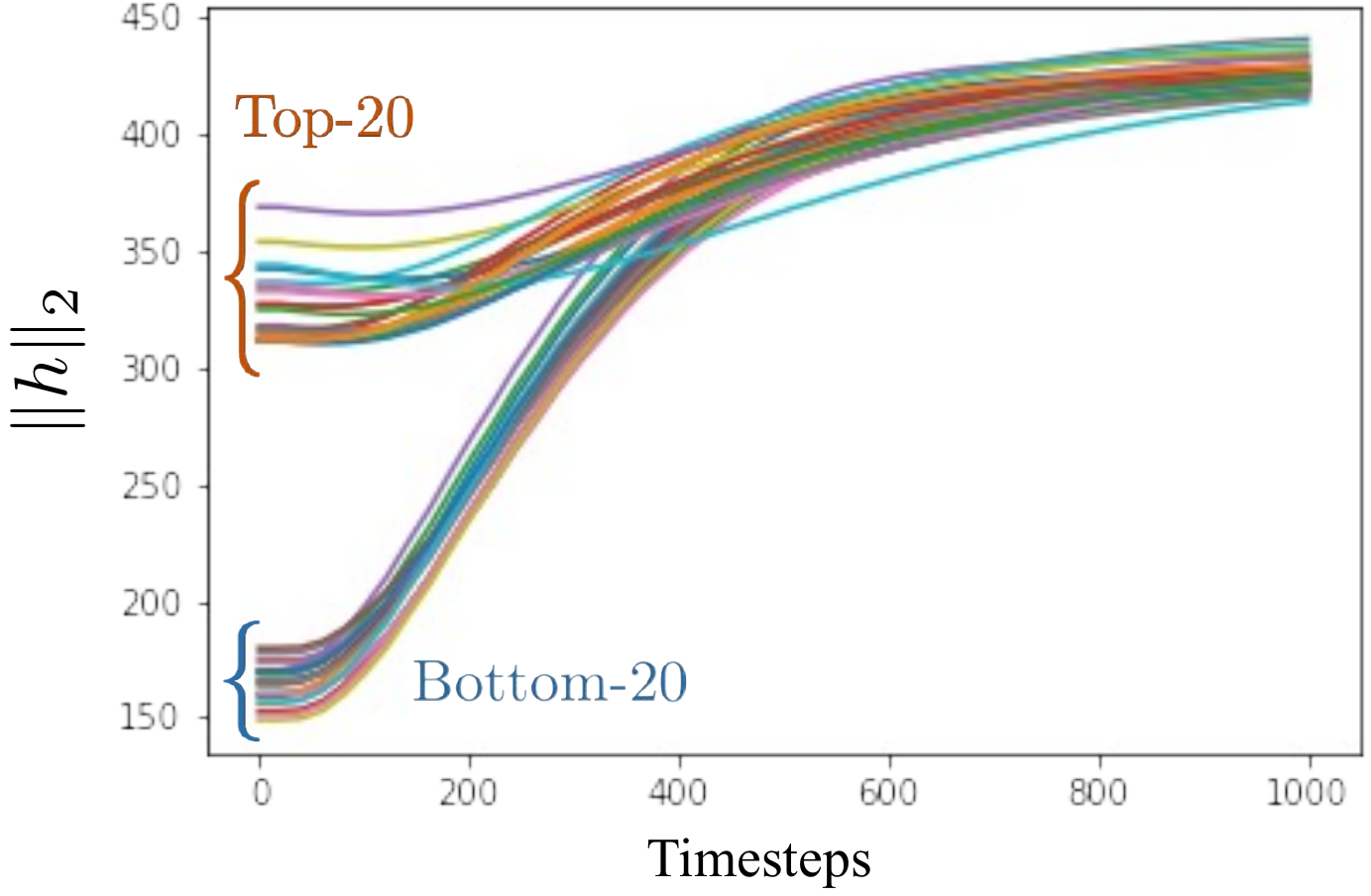}
    \vspace{-0.5em}
    \caption{We choose $\vht{}$ from the top 20 and bottom 20 samples in their norms among 500 samples. 
    Each line represents a trajectory of $\lVert h \rVert_{2}$ during the reconstruction of a sample.
}
    \vspace{-0.5em}
    \label{afig:large_gap_norms}
\end{figure}

\begin{figure}
    \centering
    \includegraphics[width=0.8\linewidth]{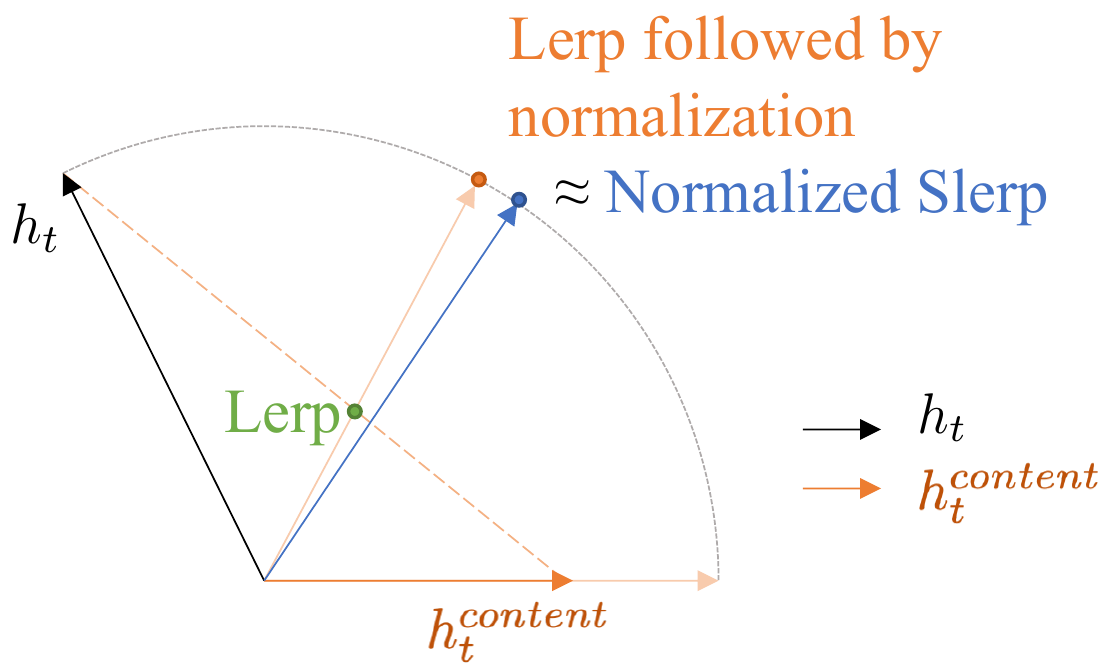}
    \vspace{-0.5em}
    \caption{Visual comparison of Slerp and Lerp. The larger difference in norms of $\vht$ and $\vhtcon$ leads to a larger gap between the results. Lerp followed by normalization is closer to Slerp than Lerp.
}
    \vspace{-0.5em}
    \label{afig:norm_visualize}
\end{figure}

\section{More analyses of Slerp}
\subsection{Comparison with Lerp}
\label{supp:lerp}

The intuition behind using Slerp is that we should preserve the correlation between $\vht$ and its matching skip connection (\sref{sec:slerp}). Here, we explore an alternative: Lerp.
When $\vht$ and $\vhtcon$ have different norms, using Lerp results in more artifacts in the final image as shown in \fref{afig:compare_lerp_slerp}. This difference in norms of $\vht$ is reported in \fref{afig:large_gap_norms}.
\fref{afig:norm_visualize} illustrates the difference between Slerp, Lerp, and Lerp followed by normalization. Lerp may change the norm of $\textbf{f}(\vht, \vhtcon, \gamma)$ when the norm of $\vht$ and $\vhtcon$ are different, leading to a decrease in image quality. However, Lerp followed by normalization produces results similar to Slerp. Still, we choose Slerp because it is easier to implement and less prone to errors.


\subsection{Cumulative content injection}
\label{supp:cumulative}
In addition to improving the quality of images, our approach allows us to control the amount of content injection by adjusting the $\vht\text{-to-}\vhtcon$ ratio through Slerp parameter $\gamma_t$. A small $\gamma_t$ results in a smaller amount of content injection. As mentioned in \sref{sec:replace}, preserving the $\vht$ component improves quality. However, there is a trade-off between the content injection rate and quality, and therefore, the value of $\vht$ needs to be constrained. Further experiments to determine the proper range of $\gamma{}$ are discussed in \sref{sec:gamma}.
 
 Note that the effects of Slerp are cumulative along the reverse process as the content injection at $t$ affects the following reverse process in $[t-1,\tedit]$.
We provide an approximation of the total amount of injected content as follows. Assuming that the angle between $\vht$ and $\vhtcon$ is close to 0 and the results of content injection at $t$ are directly passed to the next \thspace{} at $t-1$ without any loss, then
$$\tildeh_{t} = (1-\gamma) \vht + \gamma \vhtcon \approx{} f(\vht, \vhtcon, \gamma)$$
and $$ \vh_{t-1} \approx{} \tildeh_t.$$
Along the reverse process, $\tildeh_t$ is recursively fed into the next stage.
After $n$ content injections, we get 

\begin{equation}
\label{eq:cumulative}
\tildeh_{t-n} \approx{} (1-\gamma)^n \vh_t + \gamma \sum_{i=1}^{n} (1-\gamma)^{i-1} h_{t-i}^{content}.
\end{equation}
As $0 \leq \gamma \leq 1$, the proportion of $\vht$ decreases exponentially and the proportion of $\vhtcon$ accumulates during the content injection stage. It indicates that a large proportion of content is injected compared to $\gamma{}$ of Slerp. For further details regarding the ablation study on $\gamma{}$, please refer to \sref{sec:gamma}.


\begin{figure}[!t]
    \centering
    \includegraphics[width=0.9\linewidth]{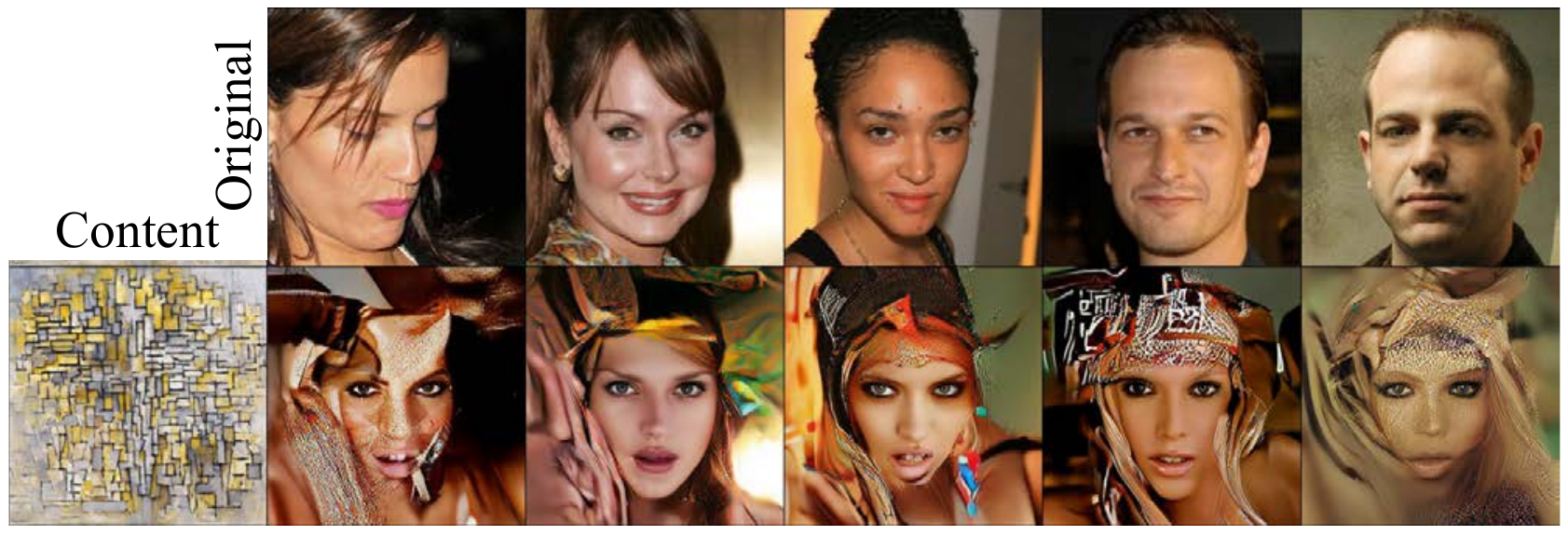}
    \caption{\textbf{Content image from unseen domain} Other than original images, $\vhtcon{}$ obtained from unseen domain results in poor images.}
    \label{fig:limitation_ood}
\end{figure}


\section{Discussion details}
\label{appendix:limitations}

As mentioned in \sref{sec:conclusion}, \fref{fig:limitation_ood} shows that using out-of-domain images as content leads to completely distorted results. It implies that $\vht$ cannot be considered a universal representation for all types of content. 

\fref{fig:limitation_mask} shows the local mixing with various feature map mask sizes. Using the feature map mask, we can designate the specific area where the content injection is applied. 
Unfortunately, the \thspace{} has small spatial dimensions, limiting the resolution of the mask for local mixing.

\begin{figure}[t]
    \centering
    \includegraphics[width=0.6\linewidth]{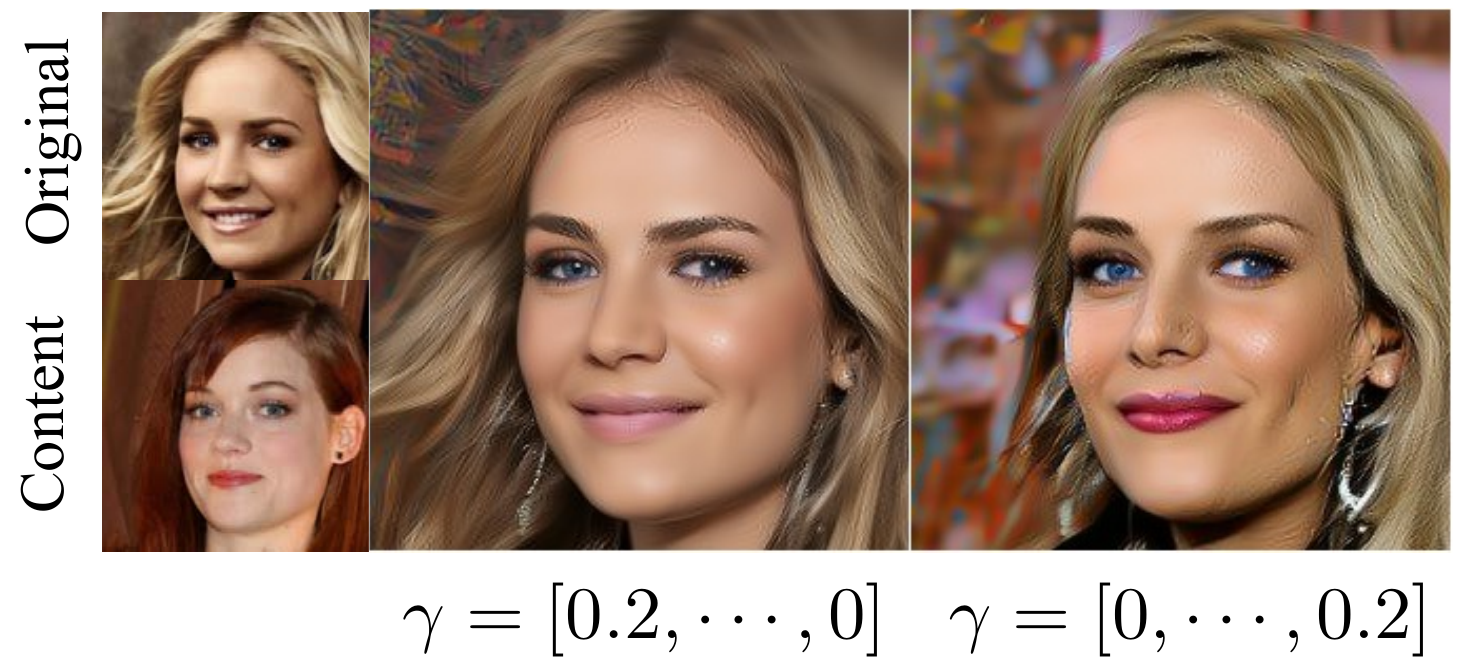}
    \vspace{-1.0em}
    \caption{\textbf{Various interpolation ratio schedule.} $\gamma{}$ is content injection rate.}
    \label{fig:rebuttal}
\end{figure}

\section{$\gamma{}$ scheduling}
\label{appendix:gamma_scheduling}

\fref{fig:rebuttal} provides the results from alternative schedules. 
Gradually decreasing the injection along the generative process enhances realism, however, it may not accurately represent the content.
Conversely, gradually increasing the injection better preserves the content but results in more artifacts. We keep the total amount of injection fixed in this experiment.


\section{More related work}
After \cite{ho2020denoising,song2020score} proposed a universal approach for Diffuson models (DMs), subsequent works have focused on controlling the generative process of DMs ~\cite{zhang2023adding,parmar2023zero,li2023gligen,couairon2022diffedit,gal2022image,yang2022paint,kumari2023multi,xie2022smartbrush,choi2021ilvr,meng2021sdedit,avrahami2022blended,mokady2022null,kim2021diffusionclip,wallace2022edict}. Especially, \cite{park2023unsupervised,kwon2022diffusion,zhu2023boundary,tumanyan2023plug,baranchuk2021label} have uncovered the role of intermediate feature maps of diffusion models and utilized it for image editing, segmentation, and translation. However, we are the first to analyze the role of the latent variables $\vx_t$ in DMs and apply it to content injection.

The research on controlling the generative process has been done in other generative models such as GANs~\cite{goodfellow2020generative}. \cite{gatys2015neural,isola2017image} introduce style transfer and image-to-image translation with GANs and there have been a number of works that focused on the style of images ~\cite{hoffman2018cycada,choi2020stargan,choi2018stargan,yoo2019photorealistic,baek2021rethinking,park2019semantic,wang2018high}. After StyleGAN~\cite{karras2019style,karras2018progressive,karras2020analyzing}, more diverse methodologies have been proposed~\cite{kim2021exploiting,karras2020analyzing,choi2018stargan,kim2021exploiting,chong2022jojogan}. However, most of them require training.


\begin{figure*}
    \centering
    \includegraphics[width=0.8\linewidth]{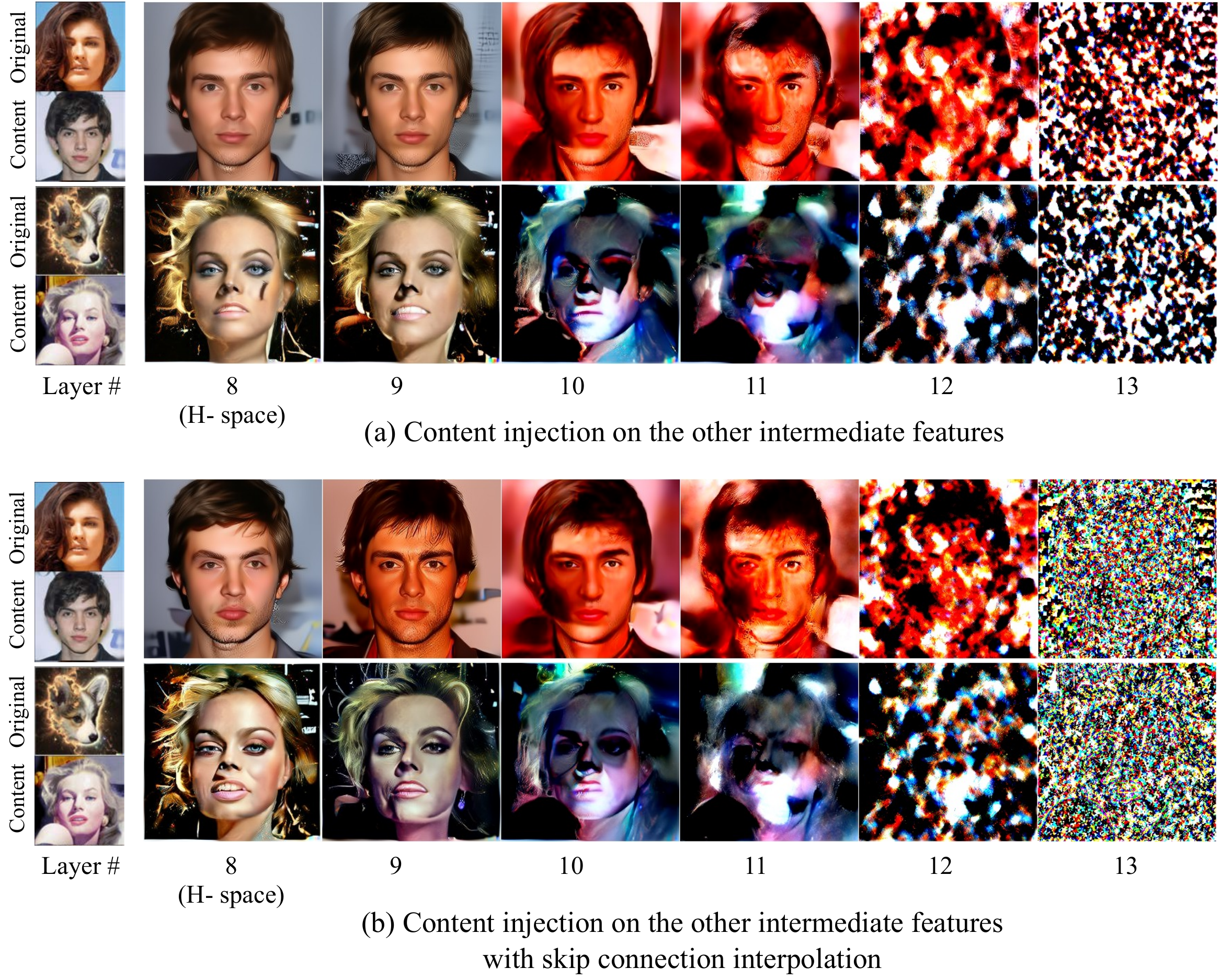}
    \captionof{figure}{
    The importance of h-space. When we inject features into additional layers, the results are disrupted. It supports h-space has semantic information and is the reason why we inject features into only h-space.
    }
    \label{fig:content_injection_on_other_layer}
    \vspace{-1em}
\end{figure*}

\section{Stable diffusion experiment details}
\label{supple:stable}

We provide more details of experiments with Stable diffusion. In \fref{fig:stablediffusion}, we use conditional random sampling with Stable diffusion v2.  In order to apply \ours{} on Stable diffusion, there are 3 options with conditional guidance. 1) content injection only with unconditional output, 2) content injection only with conditional output, 3) content injection with both conditional/unconditional outputs. 
We find that using only the unconditional output for content injection resulted in poor outcomes, while the other two options produced similar results. Thus, we use only the conditional output for content injection in \fref{fig:stablediffusion}.

Moving on to the implementation details for Stable diffusion, we set the scale to 9.0, use 50 steps for DDIM sampling, and employ the following prompts: for an original image, ``a highly detailed epic cinematic concept art CG render digital painting artwork: dieselpunk steaming robot" and for a content image: ``digital painting artwork: a cube-shaped robot with big wheels", for an original image: ``8k, wallpaper car" and for a content image: ``concept, 8k, wallpaper sports car, ferrari bg", for an original image: ``a realistic photo of a woman." and 
for a content image, ``a realistic photo of a muscle man.", original image: ``A digital illustration of a small town, 4k, detailed, animation, fantasy" and for an original image: ``A digital illustration of a dense forest, trending in artstation, 4k, fantasy."

\section{Definition of content}
\label{supple:moredetail}
We provide more details of content definition used in \sref{sec:analyses}. 
We classify each of the attributes to determine whether they are from the content image or the original image by CLIP score (CS);
\begin{equation}
\label{eq1}
\text{CLIPScore}(x,a) = 100 * \text{sim}(\mathbf{E}_{\mathbf{I}}(x),\mathbf{E}_{\mathbf{T}}(a)),
\end{equation}
where $x$ is a single image, $a$ is a given text of attribute, $\text{sim}(*,*)$ is cosine similarity, and $\mathbf{E}_{\mathbf{I}}$ and $\mathbf{E}_{\mathbf{T}}$ are CLIP image encoder and text encoder respectively.

First, we calculate the CS between the desired texts and images, original image $x_o$, content image $x_c$, and result image $x_r$. Then, if the $|\text{CS}(x_o,a)-\text{CS}(x_r,a)| > |\text{CS}(x_c,a)-\text{CS}(x_r,a)|$ then we regard the attribute is from the content image and vice versa.

In order to ignore the case that $x_o$ and $x_c$ have similar attributes, the classified result was ignored when the difference between the two values was very small. Formally, if $\| |\text{CS}(x_o,a)-\text{CS}(x_r,a)| - |\text{CS}(x_c,a)-\text{CS}(x_r,a)| \| < \lambda_{th}$, we pass that sample for that attribute. We use 5k images and set $\lambda_{th}=0.2$.

The result shows that content includes glasses, square jaw, young, bald, big nose, and facial expressions and the remaining elements include hairstyle, hair color, bang hair, accessories, beard, and makeup.

For the user study, we show the resulting image and ask people to choose the content or original image for each attribute.
We use randomly chosen 100 images and aggregate the responses from 50 participants.



\clearpage


\end{document}